\documentclass[10pt,twocolumn,letterpaper]{article}

\usepackage{iccv}
\usepackage{times}
\usepackage{epsfig}
\usepackage{graphicx}
\usepackage{amsmath}
\usepackage{amssymb}
\usepackage{subfigure}
\usepackage{booktabs}
\usepackage{multicol}
\usepackage{multirow}
\usepackage{listings}
\usepackage{appendix}
\usepackage{xcolor}

\definecolor{codegreen}{rgb}{0,0.6,0}
\definecolor{codegray}{rgb}{0.5,0.5,0.5}
\definecolor{codepurple}{rgb}{0.58,0,0.82}
\definecolor{backcolour}{rgb}{0.95,0.95,0.92}

\lstdefinestyle{mystyle}{
	backgroundcolor=\color{backcolour},   
	commentstyle=\color{codegreen},
	keywordstyle=\color{magenta},
	numberstyle=\tiny\color{codegray},
	stringstyle=\color{codepurple},
	basicstyle=\rmfamily\scriptsize,
	breakatwhitespace=false,         
	breaklines=true,                 
	captionpos=b,                    
	keepspaces=true,                 
	numbers=left,                    
	numbersep=4pt,                  
	showspaces=false,                
	showstringspaces=false,
	showtabs=false,                  
	tabsize=1
}
\usepackage{graphicx}
\usepackage{amsmath}
\usepackage{amssymb}
\usepackage{soul}
\usepackage{booktabs}
\usepackage{multirow}
\usepackage[breaklinks=true,bookmarks=false]{hyperref}

\iccvfinalcopy 

\def\iccvPaperID{****} 

\ificcvfinal\fi

\begin{document}

\title{UMC: A Unified Bandwidth-efficient and Multi-resolution based\\Collaborative Perception Framework}
\author{Tianhang Wang\textsuperscript{1}, Guang Chen\textsuperscript{1,}\thanks{Corresponding author: guangchen@tongji.edu.cn}, Kai Chen\textsuperscript{1}, Zhengfa Liu\textsuperscript{1}, Bo Zhang\textsuperscript{2}, Alois Knoll\textsuperscript{3}, Changjun Jiang\textsuperscript{1}\\
\textsuperscript{1}Tongji University, \textsuperscript{2}Westwell lab, \textsuperscript{3}Technische Universität München\\
{\tt\small \{tianya\_wang,guangchen,14sdck,cjjiang\}@tongji.edu.cn}\\
{\tt\small zhengfaliu2011@163.com,\quad zhangaigh@gmail.com, \quad knoll@in.tum.de}}


\maketitle
\ificcvfinal\thispagestyle{empty}\fi

\begin{abstract}
	Multi-agent collaborative perception (MCP) has recently attracted much attention. It includes three key processes: communication for sharing, collaboration for integration, and reconstruction for different downstream tasks. Existing methods pursue designing the collaboration process alone, ignoring their intrinsic interactions and resulting in suboptimal performance. In contrast, we aim to propose a \textbf{U}nified \textbf{C}ollaborative perception framework named UMC, optimizing the communication, collaboration, and reconstruction processes with the \textbf{M}ulti-resolution technique. The communication introduces a novel trainable multi-resolution and selective-region (MRSR) mechanism, achieving higher quality and lower bandwidth. Then, a graph-based collaboration is proposed, conducting on each resolution to adapt the MRSR. Finally, the reconstruction integrates the multi-resolution collaborative features for downstream tasks. Since the general metric can not reflect the performance enhancement brought by MCP systematically, we introduce a brand-new evaluation metric that evaluates the MCP from different perspectives. To verify our algorithm, we conducted experiments on the V2X-Sim and OPV2V datasets. Our quantitative and qualitative experiments prove that the proposed UMC greatly outperforms the state-of-the-art collaborative perception approaches.

\end{abstract}

\section{Introduction}
Single-vehicle perception has made remarkable achievements in object detection\cite{Lin2017FeaturePN, Redmon2016YouOL, Ren2015FasterRT}, segmentation\cite{Qi2017PointNetDL, ronneberger2015u}, and other tasks with the advent of deep learning. However, single-vehicle perception often suffers from environmental conditions such as occlusion\cite{Wang2020RobustOD, Yuan2021RobustIS} and severe weather\cite{Chen2021ALLSR, Li2020AllIO, Zhang2021DeepDM}, making accurate recognition challenging. To overcome these issues, several appealing studies have been devoted to collaborative perception\cite{Chen2019FcooperFB, Chen2019CooperCP, coopernaut, Wu2020jun, yu2022dairv2x}, which take advantage of sharing the multiple-viewpoint of the same scene with the Vehicle-to-Vehicle(V2V) communication\cite{9001049}. 

\begin{figure}[t]
	\centering
	\includegraphics[width=\linewidth]{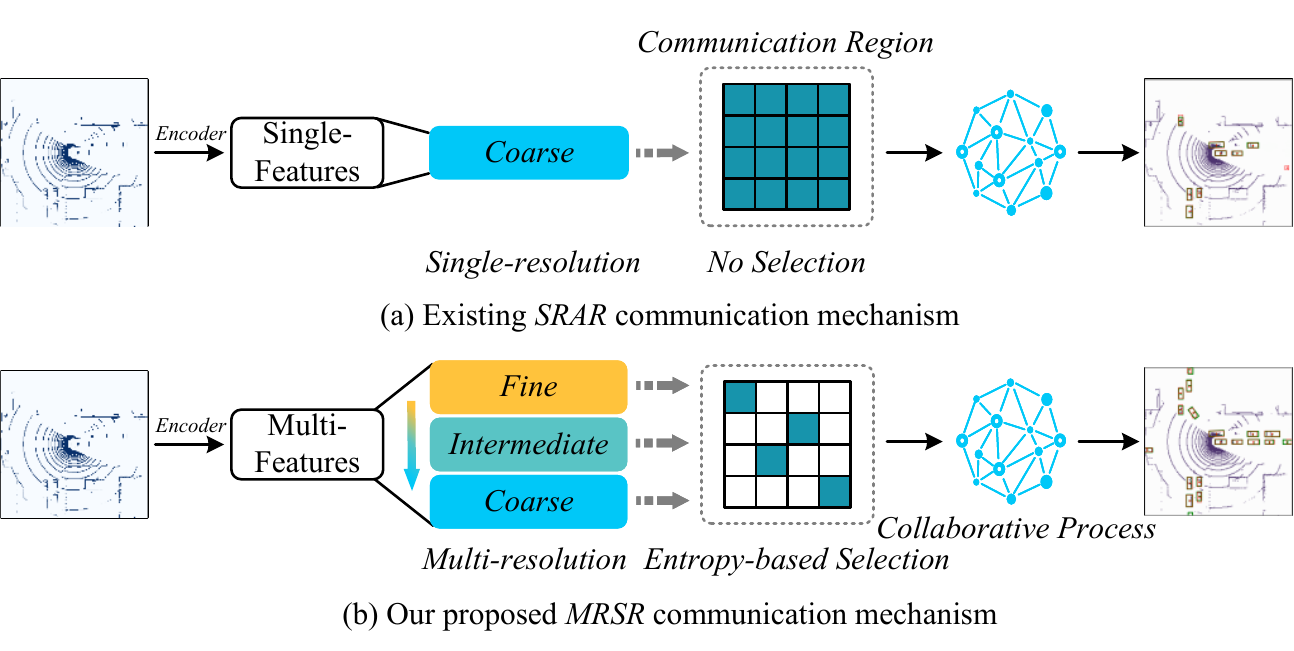}
	\caption{\textbf{The SRAR versus the proposed MRSR mechanism}. (a) The existing single-resolution and all-region (SRAR) method utilizes the coarse-grained feature with bandwidth inefficient all-region transmission. (b) The proposed MRSR mechanism incorporates bandwidth-efficient entropy-based selection with multi-resolution features.}
\label{fig:diffdd}
\end{figure}

To design a collaborative perception algorithm, current approaches\cite{Wang2020V2VNetVC, Liu2020Who2comCP, Xu2022V2XViTVC, Li2021LearningDC, xu2022cobevt} mainly focus on the collaboration process alone, aiming to design a high-performance collaboration strategy. Nevertheless, such a strategy ignores intrinsic interactions with two critical processes: communication\cite{Liu2020When2comMP,Where2comm:22} and reconstruction\cite{He2018DeepSF}. 
This kind of collaboration strategy will inherently cause suboptimal performance because the quality of communication directly determines the performance of collaboration\cite{Li2021LearningDC}, and the reconstruction influences the quality of feature maps for downstream tasks. Meanwhile, the collaboration can also affect the design of communication and reconstruction. Hence, failing to optimize these three processes will degrade the system.

In this paper, to the best of our knowledge, we are the first to propose a unified collaborative perception framework that optimizes the communication, collaboration, and reconstruction processes with multi-resolution technique. As for communication, as shown in Figure~\ref{fig:diffdd}, we propose a novel multi-resolution and selective-region (MRSR) mechanism, which is different from the single-resolution and all-region (SRAR) mechanism that is used by existing collaborative algorithms\cite{Li2021LearningDC,Xu2022V2XViTVC,Wang2020V2VNetVC}. In MRSR, we replace the coarse-grained feature in SRAR with multi-grain features, which reflect the scene from a global to a local perspective. The multi-grain features will have the complementary advantages of global structure information and local texture details. Meanwhile, to reduce the burden of bandwidth, unlike SRAR, which blindly transmits all regions, we implement a trainable and region-wise entropy-based communication selection (Entropy-CS) module that highlights the informative regions and selects appropriate regions for each level of resolution. The proposed MRSR is adaptive to real-time communication, reflecting dynamic connections among agents. 

We propose a graph-based collaborative GRU(G-CGRU) module for MRSR's each level resolution feature in collaboration. Compared to the existing collaboration strategies that only focus on the present shared information without considering the previous state, we redesign the GRU-cell\cite{Cho2014OnTP} to adapt the time series of collaborative perception, which models the continuity of vehicle movement. In G-CGRU, the collaboration depends not only on collaborators but also on the last moment information of the ego agent. Meanwhile, we propose matrix-valued gates in G-CGRU to ensure collaboration at a high spatial resolution and allow the agents to adaptively weight the informative regions.

To adapt the proposed multi-resolution collaboration mechanism for reconstruction, we propose a multi-grain feature enhancement (MGFE) module to strengthen the feature reconstruction process. In MGFE, the fine-grained collaborative feature maps will give direct guidance to the agents’ feature maps via a global pooling operation, and the coarse-grained collaborative feature maps will again enhance the agents’ feature maps from a global perspective. This design allows the agents to comprehensively reconstruct the feature maps from local and global viewpoints.

The general evaluation metric (\textit{e.g., average precision}) can reflect the overall performance of multi-agent collaborative perception (MCP). However, the performance enhancement brought by the collaboration concept that MCP introduced cannot be reflected directly. To address this issue, we introduce a brand new evaluation metric that systematically evaluates the MCP from four aspects.

To validate the proposed framework, we conducted comprehensive experiments in 3D object detection on V2X-Sim\cite{li2022v2x} and OPV2V\cite{xu2022opencood} datasets. Our proposed unified, bandwidth-efficient and multi-resolution based collaborative perception framework (UMC) achieves a better performance-bandwidth trade-off than the state-of-the-art SRAR-based collaboration methods. Our contributions are listed as follows:

\begin{itemize}
\item We present a unified, bandwidth-efficient framework (UMC) for collaborative perception, which optimizes the communication, collaboration, and reconstruction processes with the multi-resolution technique.
\item We propose a novel multi-resolution and selective-region mechanism for communication and the graph-based collaborative GRU for each resolution collaboration and multi-grain feature enhancement module for reconstruction.
\item We present a brand new evaluation metric for 3D object collaborative perception, which can evaluate the performance from different perspectives.
\end{itemize}

\section{Related works}
\label{sec:formatting}

\subsection{V2X Collaborative Perception} F-Cooper\cite{Chen2019FcooperFB} is the first to apply feature-level collaboration, which weights interaction information equally using a max-based function. When2com\cite{Liu2020When2comMP} employs an attention mechanism to build a bandwidth-efficient communication group. V2VNet\cite{Wang2020V2VNetVC} proposes a spatially-aware message transmission mechanism that assigns varying weights to the different agents. DiscoNet\cite{Li2021LearningDC} adapts the knowledge distillation to enhance student model performance by constraining the teacher model, which is based on corresponding raw data collaboration. V2X-ViT\cite{Xu2022V2XViTVC} presents a unified V2X framework based on Transformer that takes into account the heterogeneity of V2X systems. Where2comm\cite{Where2comm:22} takes the advantage of sparsity of foreground information in detection downstream task to save the communication bandwidth of V2X collaboration. CoBEVT\cite{xu2022cobevt} generates BEV map predictions with multi-camera perception by V2X.

\subsection{3D Perception for Autonomous Driving}

Due to the declining price of commodity LiDAR sensors, 3D perception has become more prevalent in autonomous driving. VoxelNet\cite{Zhou2018VoxelNetEL} pioneered utilizing 3D object detection in autonomous driving. More and more advanced algorithms have been discovered. There are three branches of 3D perception input forms that have been widely adopted: point-based methods\cite{Qi2017PointNetDL, Shi2019PointRCNN3O, Qi2017PointNetDH} directly process raw captured point; and voxel-based methods\cite{Zhou2018VoxelNetEL, Yan2018SECONDSE, Kuang2020VoxelFPNMV, Shi2020PVRCNNPF} divide the points into a volumetric grid; and 2D projections based methods\cite{Simon2018ComplexYOLOAE, Yang2018PIXORR3, Lang2019PointPillarsFE} compact the input by projecting points into images. In UMC, we utilize bird's eye view projection as input, which decreases the processing time and enables the use of 2D object detection networks.

\begin{figure*}[t]
	\centering
	\includegraphics[width=\linewidth]{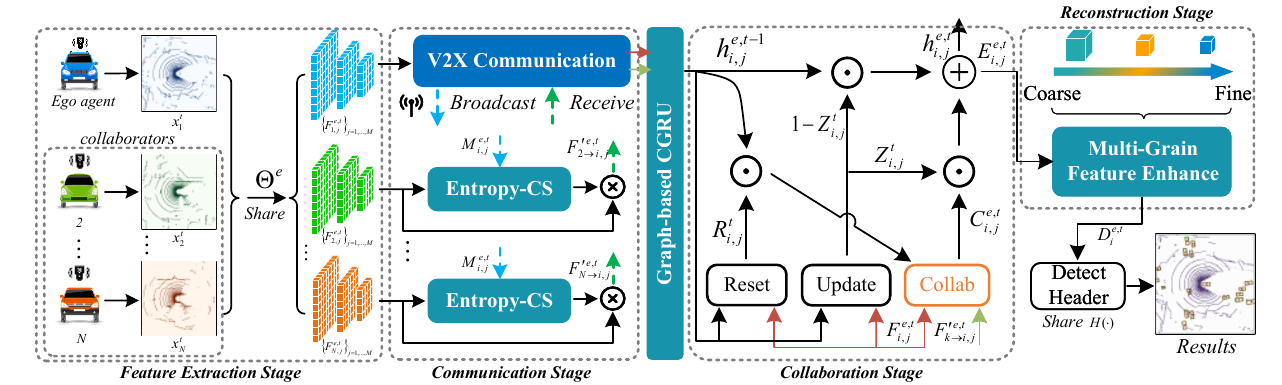}
	\caption{The overview of the proposed UMC framework. i) \textit{\textbf{Feature extraction stage}}, the agents obtain the $\boldsymbol{F}^{e,t}_{i}$ by the shared feature encoder $\Theta^{e}$ with the observation $\boldsymbol{x}^{t}_{i}$. ii) \textit{\textbf{Communication stage}}, the ego agent (in \textbf{\textcolor{blue}{blue}}) will broadcast compact query matrix $\boldsymbol{M}^{e,t}_{i}$ in each resolution to collaborators by V2X communication, and the collaborators will compute the transmission map by entropy-CS module at local. Then the ego agents will receive the selected messages from the assisted collaborators. iii) \textit{\textbf{Collaboration stage}}, the ego agents will employ the G-CGRU module in each resolution for high efficient collaboration. iv) \textit{\textbf{Reconstruction stage}}, the MGFE module will reconstruct the ego agent's feature by multi-resolution collaborative feature maps for different downstream tasks.}
	\label{fig:framework}
\end{figure*}

\subsection{Multi-resolution Image Inpainting}

Multi-resolution is a widely used image inpainting\cite{9577298, 9578341, Wang2021ParallelMF} technique to combine low-resolution feature maps that contain global structure information with high-resolution feature maps that are rich in local texture details, such as, PEN-Net\cite{yan2019PENnet} proposes the attention map mechanism, which learns from high-resolution feature maps to guide low-resolution features. Wang \etal~\cite{Wang2021ParallelMF} design a parallel multi-resolution fusion network that can effectively fuse structure and texture information. DFNet\cite{10.1145/3343031.3351002} employs U-Net\cite{Ronneberger2015UNetCN} based multiple fusion blocks to implement multi-resolution constraints for inpainting. In UMC, we implement the multi-resolution technique for a new application scenario: multi-agent collaborative perception. The multi-grain collaborative feature makes it possible for the agents to reconstruct the occlusion area from both a local and a global point of view.

\section{UMC}

In this section, we present the technical details of the proposed UMC. The overall architecture of the proposed UMC is shown in Figure~\ref{fig:framework}, which consists of three new modules, entropy-CS, G-CGRU, and the MGFE. 
Entropy-CS extends the traditional entropy theory to select the informative regions of observations, reducing the heavy bandwidth burden brought by multi-resolution.
G-CGRU redesigns the GRU-Cell to adapt the multi-agent collaboration, which models the continuity of vehicle movement. MGFE introduces the multi-grain feature enhancement to strengthen the feature reconstruction process for different downstream tasks.

\subsection{Problem Statement and Notation}

We assume there are $ N $ collaborators with their corresponding observations $\boldsymbol{X}^{t} = \{\mathit{x}^{t}_{n}\}_{n=1,\dots,N}$ at time $t$. Among those collaborators, the feature set of the $i$-th agent is defined as $\boldsymbol{F}^{e,t}_{i}=\{\boldsymbol{F}^{e,t}_{i,j}\}_{j=1,\dots,M} \gets \Theta^{e}(x_{i,t})$, where $\Theta(\cdot)$ is the feature encoder shared by all the collaborators and $ M $ represents the number of intermediate layers in $\Theta(\cdot)$. 

Note that because each agent has its unique pose $\boldsymbol{\xi}^{t}$, we need conduct feature alignment operation $\Lambda(\cdot,\cdot)$ across agents for subsequent collaboration process, \eg, for the $i$-th agent, the transformed feature map from the $k$-th agent of the $j$-th intermediate layer is $\boldsymbol{F}^{e,t}_{k \to i, j} \gets \Lambda(\boldsymbol{F}^{e,t}_{k,j}, \boldsymbol{\xi}^{t}_{k \to i})$, where the $\boldsymbol{\xi}^{t}_{k \to i}$ is based on two ego poses $\boldsymbol{\xi}^{t}_{k}$ and $\boldsymbol{\xi}^{t}_{i}$. Now the $\boldsymbol{F}^{e,t}_{k \to i, j}$ and $\boldsymbol{F}^{e,t}_{i, j}$ are supported in the same coordinate system. The goal of this paper is to optimize the unified collaborative perception framework and each agent outputs the prediction of perception task: \resizebox{0.47\textwidth}{!}{
	$\hat{\boldsymbol{Y}}^{t} = \sum^{N}_{i=1}\{\hat{\boldsymbol{y}}^{t}_{i}|\boldsymbol{H}(\Psi(\sum^{M}_{j=1}\Upsilon(\boldsymbol{F}^{e,t}_{i,j},\sum^{N}_{k \ne  i}\Omega(\boldsymbol{F}^{e,t}_{k \to i,j}))))\}$}, where $\Omega(\cdot), \Upsilon(\cdot,\cdot), \Psi(\cdot)$ represent the entropy-CS, G-GGRU, and MGFE module, respectively. The $\boldsymbol{H}(\cdot)$ denotes the headers for different perception tasks.

\subsection{Entropy-based Communication Selection}
\begin{figure}[t]
	\centering
	\includegraphics[width=\linewidth]{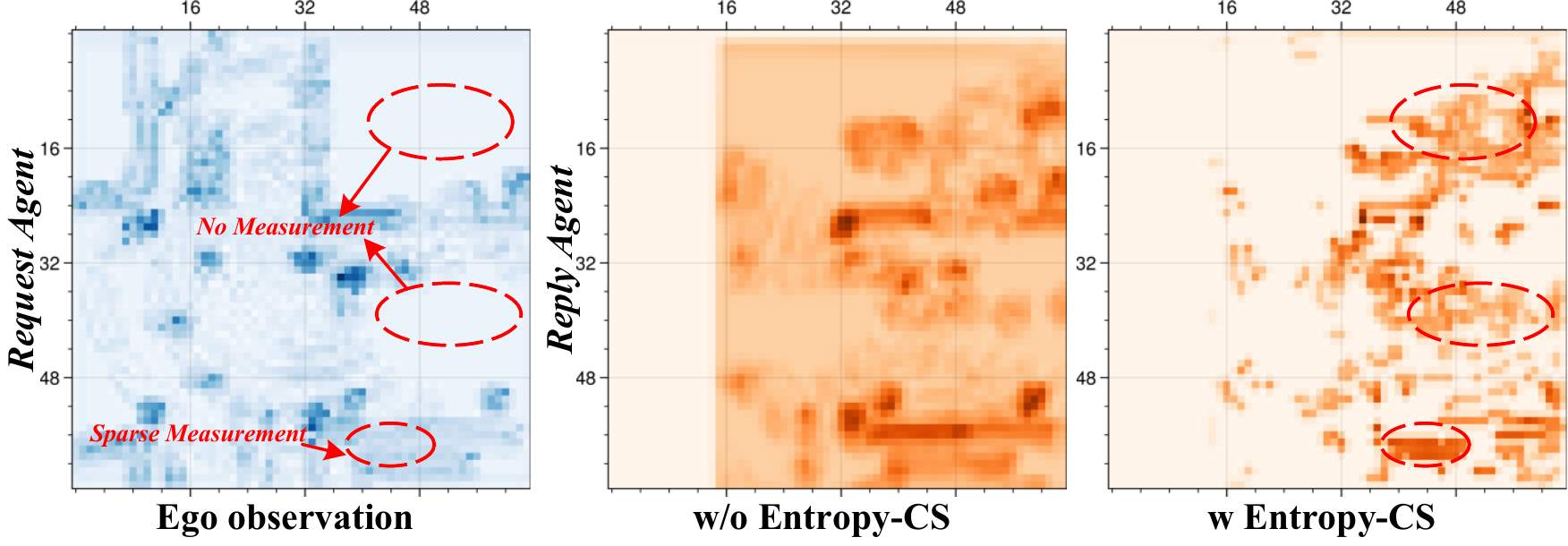}
	\caption{Illustration of entropy-based communication selection. The entropy-CS filters out the lower quality regions, achieving high efficient bandwidth usage.}
\label{fig:entropy}
\end{figure}

The intuition of entropy-CS is low computational complexity and high interpretability to filter out the lower-quality regions. Based on above considerations, we extend the traditional entropy theory\cite{crooks2017measures} toward two-dimensional communication selection. Hence, the entropy-CS is non-parameter and high efficient to reduce the heavy bandwidth burden brought by multi-resolution feature information. 

To further compress bandwidth requirements, the ego agent $i$ and collaborator will first compress their encoder features $\boldsymbol{F}^{e,t}_{i,j}, \boldsymbol{F}^{e,t}_{k \to i,j}  \in \mathbb{R}^{K,K,C}$ along channels into the query matrix $\boldsymbol{M}^{e,t}_{i,j}, \boldsymbol{M}^{e,t}_{k \to i,j} \in \mathbb{R}^{K,K}$ through a trainable $1\times1$ convolution kernel $\boldsymbol{W}_{1 \times 1}$, which is supervised by the downstream loss function. Then, we construct the two-stage entropy-based selective-region communication mechanism:

\underline{\textbf{\textit{Self-select stage}}} We first let the $k$-th collaborator determine location $\boldsymbol{S}^{e,t}_{k \to k,j}$ itself, where potentially contains useful information for later collaboration:
\begin{equation}
\setlength{\abovedisplayskip}{3pt}
\setlength{\belowdisplayskip}{3pt}
\label{eq:self}
\boldsymbol{S}^{e,t}_{k \to k,j} = \boldsymbol{\Gamma}(\Phi(\boldsymbol{M}^{e,t}_{k \to i,j}, \boldsymbol{M}^{e,t}_{k \to i,j}), \delta_{s})
\end{equation}
\abovedisplayshortskip=0pt
\belowdisplayshortskip=0pt
where the $\boldsymbol{\Gamma}(\cdot,\delta_{s})$ is an element-wise function, zeroing out elements smaller than the values of top-$\delta_{s}\%$ in $\cdot$, and $\Phi$ is the entropy estimation function, which measures the correlation from $\boldsymbol{K} \in \mathbb{R}^{K,K}$ to $\boldsymbol{Q} \in \mathbb{R}^{K,K}$ at the location of $\boldsymbol{L}$. Note that if $\boldsymbol{L}$ is not assigned, the $\Phi$ will compute all region of $\boldsymbol{K},\boldsymbol{Q}$:
\begin{equation}
\setlength{\abovedisplayskip}{3pt}
\setlength{\belowdisplayskip}{3pt}
\begin{aligned}
	\label{eq:entropy}
	& \Phi(\boldsymbol{K},\boldsymbol{Q}, \boldsymbol{L}) = \{p^{mn} * log(p^{mn})\}_{m, n \in \boldsymbol{L}}, \mathrm{with} \\
	& p^{mn} = \frac{1}{M*N}\sum^{M,N}_{j,k} \sigma(\boldsymbol{K}(m+j,n+k) - \boldsymbol{Q}(m,n))\nonumber
\end{aligned}
\end{equation}
where $\sigma(\cdot)$ is Sigmoid function. Note that if $\Arrowvert \boldsymbol{S}^{e,t}_{k \to k,j}\Arrowvert \simeq 0$, represents that the $k$-th collaborator almost has no sufficient information. The entropy-CS will close the later cross-select communication stage, which can further reduce the bandwidth requirements.

\underline{\textbf{\textit{Cross-select stage}}} After the self-entropy selection stage, we thus derive the cross-entropy selection to obtain the communication location $\boldsymbol{S}^{e,t}_{k \to i,j}$ with the broadcast  $\boldsymbol{M}^{e,t}_{i,j}$:
\abovedisplayshortskip=0pt
\belowdisplayshortskip=0pt
\begin{equation}
\setlength{\abovedisplayskip}{3pt}
\setlength{\belowdisplayskip}{3pt}
\label{eq:self1}
\boldsymbol{S}^{e,t}_{k \to i,j} = \boldsymbol{\Gamma}(\Phi(\boldsymbol{M}^{e,t}_{i,j}, \boldsymbol{M}^{e,t}_{k \to i,j}, \boldsymbol{S}^{e,t}_{k \to k,j}), \delta_{c})
\end{equation}

As demonstrated in Figure~\ref{fig:entropy}, the entries of the derived matrix $\boldsymbol{S}^{e,t}_{k \to i,j}$ indicate where to communicate, which requires about $\frac{1}{\delta_{s}\delta_{c}}$ bandwidth of existing SRAR-based methods\cite{Wang2020V2VNetVC,Li2021LearningDC, Liu2020When2comMP}. However, the transmission $\boldsymbol{T}^{e,t}_{k\to i,j} \gets \boldsymbol{F}^{e,t}_{k \to i,j}[\boldsymbol{S}^{e,t}_{k \to i,j}]$ is sparse, with many positions being $\boldsymbol{0}$. This could potentially harm further collaboration process\cite{liu2022anytime}. To compensate for this, after the ego agent receives, we spatially interpolate these positions from their neighbors across all channels at local, using a similar approach in \cite{xie2020spatially}: $\boldsymbol{F}^{'e,t}_{k \to i,j} \gets Interpolate(\boldsymbol{T}^{e,t}_{k\to i,j})$.

\subsection{Graph-based Collaborative GRU}

Once a requesting ego agent collects the information from its linked supporting collaborators, it is important to design how to integrate its local observation with the selected multi-resolution feature maps from supporters\cite{Li2021LearningDC}.

We observe that the detected agents in perception area change slowly over time in the urban low-speed collaborative scenes\cite{xu2022opencood, li2022v2x}. Hence, the collaborative feature maps from last time are actually helpful to the present process.

However, the general RNN model\cite{rumelhart1986learning, hochreiter1997long} can only process one-dimensional features, which makes it hard to preserve the spatial structure and local details of the agent's observation.

Based on the above considerations, we propose a novel graph-based collaborative GRU module named G-CGRU for each resolution collaboration process, as demonstrated in Figure~\ref{fig:framework}. The key intuition of G-CGRU is that the collaboration depends not only on supporters but also on requesting the ego agent's previous information.

For the ego agent $i$ of the $j$-th resolution intermediate feature maps, the inputs of G-CGRU are hidden states $\boldsymbol{h}^{e,t-1}_{i,j} \in \mathbb{R}^{K,K,C}$, the ego agent observation $\boldsymbol{F}^{e,t}_{i,j}$, and the supporters' selected feature maps $\{\boldsymbol{F}^{'e,t}_{k \to i,j}\}_{k \ne i}$. The module mainly has three key gates as follows:

\underline{\textbf{\textit{Update and Reset gates}}} The $Update$ gate is responsible for determining where the previous information needs to be passed along the next state and the $Reset$ is to decide how much of the past information is needed to neglect. The $Update$ and $Reset$ gates share the same structure but different weights. Here we take $Reset$ as an examples:
\abovedisplayshortskip=0pt
\belowdisplayshortskip=0pt
\begin{equation}
\setlength{\abovedisplayskip}{3pt}
\setlength{\belowdisplayskip}{3pt}
\begin{aligned}
	& \boldsymbol{W}_{ir} = \sigma(\boldsymbol{W}_{3 \times 3}*([\boldsymbol{h}^{e,t-1}_{i,j};\boldsymbol{F}^{e,t}_{i,j}])) \\
	& \boldsymbol{R}^{t}_{i,j} = \sigma(\boldsymbol{W}_{ir} \odot \boldsymbol{\hat{h}}^{e,t-1}_{i,j} + (1 - \boldsymbol{W}_{ir}) \odot \boldsymbol{F}^{e,t}_{i,j})
\end{aligned}
\label{eq:reset}
\end{equation}
\noindent where $\odot, \sigma(\cdot), [\cdot;\cdot]$ represent dot product, Sigmoid function, and concatenation operation along channel dimensions, respectively. $*$ indicate a $3\times3$ convolution operation.
The $\boldsymbol{R}^{t}_{i,j} \in \mathbb{R}^{K,K,C}$ learns where the hidden features $\boldsymbol{h}^{e,t-1}_{i,j}$ are conducive to the present.

\underline{\textbf{\textit{Collab gate}}} Based on the above $Reset$ and $Update$ gates, we thus derive the $Collab$ gate. To make better collaborative feature integration, we construct a collaboration graph $\mathcal{G}^{t}_{c}(\boldsymbol{\mathcal{V}},\boldsymbol{\mathcal{E}})$ in $Collab$ gate, where node $\boldsymbol{\mathcal{V}} = \{\mathcal{V}_{i}\}_{i\in1,\dots,N}$ is the set of collaborative agents with the real-time pose
information in the environment and $\boldsymbol{\mathcal{E}} = \{\boldsymbol{W}_{i \to j}\}_{i,j\in1,\dots,N, i\ne j}$ is the set of trainable edge matrix weights between agents and models the collaboration strength between two agents. Let $\mathcal{C}_{\mathcal{G}^{t}_{c}}(\cdot)$ be the collaboration process defined in the $Collab$ module's graph $\mathcal{G}^{t}_{c}$. The $j$-th resolution enhanced maps of ego $i$ agent after collaboration are $\boldsymbol{E}^{e,t}_{i,j} \gets \mathcal{C}_{\mathcal{G}^{t}_{c}}(\boldsymbol{h}^{e,t-1}_{i,j}, \boldsymbol{F}^{'e,t}_{k \to i,j}, \boldsymbol{F}^{e,t}_{i,j})$. This process has two stages: message attention (\textbf{S1}) and message aggregation (\textbf{S2}).

In the \textit{message attention stage} (\textbf{S1}), each agent determines the matrix-valued edge weights, which reflect the strength from one agent to another at each individual cell. To determine the edge weights, we firstly get the conductive history information from hidden features by $Reset$ gates through $\boldsymbol{\hat{h}}^{e,t-1}_{i,j} \gets \boldsymbol{h}^{e,t-1}_{i,j}  \odot \boldsymbol{R}^{t}_{i,j}$. Then, we utilize the edge encode $\Pi$ to correlate the history information, the feature map from another agent and ego feature map; that is, the matrix-value edge weight from $k$-th agent to the $i$-th agent is $\boldsymbol{W}_{k \to i}=\Pi(\boldsymbol{\hat{h}}^{e,t-1}_{i,j},\boldsymbol{F}^{'e,t}_{k \to i,j},\boldsymbol{F}^{e,t}_{i,j}) \in \mathbb{R}^{K,K}$, where $\Pi$ concatenates three feature maps along the channel dimension and then utilizes four $1 \times 1$ convolutional layers to gradually reduce the number of channels from $3C$ to 1. Also, to normalize the edge weights across different agents, we implement a softmax operation on each cell of the feature map. Compared with previous work\cite{Liu2020When2comMP,Where2comm:22,Xu2022V2XViTVC} generally consider time-discrete edge weight to reflect the static collaboration strength between two agents; while we consider time-continuous edge weight $\boldsymbol{W}_{k \to i}$. which dynamically models the collaboration strength from the $k$-th agent to the $i$-th agent from $t-1$ to $t$. In the \textit{message aggregation stage} (\textbf{S2}), each agent aggregates the feature maps from collaborators based on the normalized matrix-valued edge weights, the updated feature map $\boldsymbol{C}^{e,t}_{i,j}$ is utilized by $\sum^{N}_{k=1} \boldsymbol{W}_{k \to i} \circ \boldsymbol{F}^{'e,t}_{k \to i,j}$, where $\circ$ represents the dot production broadcasting along the channel dimension. Finally, the collaborative map is $\boldsymbol{E}^{e,t}_{i,j} = \boldsymbol{Z}^{t}_{i,j} \odot \boldsymbol{C}^{e,t}_{i,j} + (1 -  \boldsymbol{Z}^{t}_{i,j}) \odot \boldsymbol{h}^{e,t-1}_{i,j}$. Note that the $\boldsymbol{Z}^{t}_{i,j}$ is generated by $Update$ gate and $\odot$ is the dot product.

\subsection{Multi-grain Feature Enhancement Module}

The final ego agent $i$ collaborative feature maps $\{\boldsymbol{E}^{e,t}_{i,j}\}_{j=1,\dots,M}$ with different resolutions generated from G-CGRU contain various visual context information, and each of them can be used to yield the downstream task. 

\begin{figure}[t]
\centering
\includegraphics[width=0.95\linewidth]{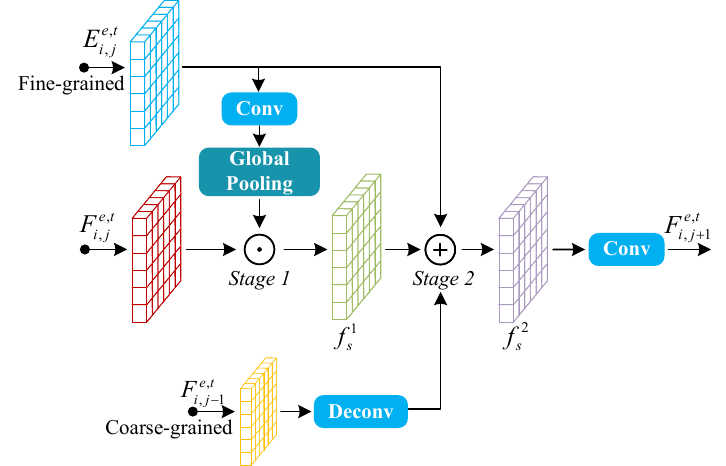}
\caption{The architecture of multi-grain feature enhancement module. The fine- and coarse-grained feature maps guide the reconstruction process from the local to global perceptive, respectively.}
\label{fig:fusion}
\end{figure}

To make the best use of multi-grain feature maps, we further propose a multi-grain feature enhancement (MGFE) module, which utilizes coarse- to fine-grained collaborative feature maps to guide the reconstruction process from a local to global perspective, as demonstrated in Figure~\ref{fig:fusion}. The MGFE consists of two stages:

\underline{\textbf{\textit{Stage one}}} A $1 \times 1$ convolution is applied to adjust the feature space of the collaborative feature map $\boldsymbol{E}^{e,t}_{i,j}$ to the ego observed feature map $\boldsymbol{F}^{e,t}_{i,j}$, then through global pooling along channels to the adjusted collaborative feature to obtain highlighted informative regions, following multiplied with the local feature map. The output $\boldsymbol{f}^{1}_{s}$ as follows:
\abovedisplayshortskip=0pt
\belowdisplayshortskip=0pt
\begin{equation}
\setlength{\abovedisplayskip}{5pt}
\setlength{\belowdisplayskip}{5pt}
\boldsymbol{f}^{1}_{s} = \mathcal{G}(\sigma(\boldsymbol{W}_{1 \times 1} * \boldsymbol{E}^{e,t}_{i,j}+\boldsymbol{b})) \circ \boldsymbol{F}^{e,t}_{i,j}
\label{eq:fusion1}
\end{equation}
\abovedisplayshortskip=0pt
\belowdisplayshortskip=0pt
\noindent where $\mathcal{G}, \circ$ denotes global max pooling operation along the channels and dot production broadcasting along the channel dimension, respectively. $\boldsymbol{W}_{1 \times 1}$ indicates trainable parameters, $\boldsymbol{b}$ refers bias.

\underline{\textbf{\textit{Stage two}}} The coarse-grained, ego-observed feature map $\boldsymbol{F}^{e,t}_{i,j-1}$ is upsampled to the same dimension as the high-resolution map, and then concatenated with the $\boldsymbol{f}^{1}_{s}, \boldsymbol{E}^{e,t}_{i,j}$. The output of stage two $\boldsymbol{f}^{2}_{s}$ as follows:
\abovedisplayshortskip=0pt
\belowdisplayshortskip=0pt
\begin{equation}
\boldsymbol{f}^{2}_{s} = [L_{2}(upsample(\boldsymbol{F}^{e,t}_{i,j-1}));L_{2}(\boldsymbol{f}^{1}_{s});L_{2}(\boldsymbol{E}^{e,t}_{i,j})]
\label{eq:fusion2}
\end{equation}
\noindent where $upsample$ denotes the deconvolution operation, $L_{2}$ refers to the $L_2$ norm layer\cite{Liu2015ParseNetLW}, which is helpful for combining two (or more) feature maps. To adjust the channels dimension of $\boldsymbol{f}^{2}_{s}$, a $3 \times 3$ convolution is applied to $\boldsymbol{f}^{2}_{s}$: $\boldsymbol{F}^{e,t}_{i,j+1} \gets \sigma(\boldsymbol{W}_{3 \times 3}*\boldsymbol{f}^{2}_{s} + \boldsymbol{b})$.

After the reconstruction process of the MGFE module, we can obtain the ego agent enhancement feature $\boldsymbol{D}^{e,t}_{i}$. To obtain the final detection outputs $\boldsymbol{\tilde{Y}^{t}_{i}}$, we employ an output header $\boldsymbol{H}(\cdot)$: $\boldsymbol{\tilde{Y}^{t}_{i}} \gets \boldsymbol{H}(\boldsymbol{D}^{e,t}_{i})$. To implement the header $\boldsymbol{H}(\cdot)$, we employ two branches of convolutional layers to classify the foreground-background categories and regress the bounding boxes.

To train our model, we employ the binary cross-entropy loss to supervise foreground-background classification and the smooth $L_{1}$ loss to supervise the bounding-box regression:
\abovedisplayshortskip=0pt
\belowdisplayshortskip=0pt
\begin{equation}
\mathcal{L} = \sum^{N}_{i=1}\mathcal{L}_{det}(\boldsymbol{Y}_{i}, \boldsymbol{\tilde{Y}}_{i}) 
\label{eq:loss}
\end{equation}

\noindent where $\mathcal{L}_{det}$ denotes the classification and regression losses and $\boldsymbol{Y}_{i}, \boldsymbol{\tilde{Y}_{i}} $ represents the ground-truth detection and predicted detection in the perception region of the $i$-th agent, respectively.

\section{Experiment}

\begin{table*}[t]
\centering%
\begin{minipage}[b]{0.48\linewidth}
	\centering%
	\caption{Detection comparison on V2X-Sim dataset. Key: [\textbf{\textcolor{red}{Best}}, \textbf{\textcolor{blue}{Second Best}}, \textbf{\textcolor{green}{Third Best}}]}
	\label{table:v2x-sim}
	\resizebox{1.0\textwidth}{!}{
		\begin{tabular}{clllll}
			\hline
			Models     & $\mathrm{ARSV}_{50/70}\uparrow$ & $\mathrm{ARCV}_{50/70} \uparrow$ & $\mathrm{ARCI}_{50/70}\downarrow$ &$\mathrm{ARTC}_{50/70}\uparrow$ & $\ \mathrm{AP}_{50/70} \uparrow $          \\ \hline
			No Fusion  &      76.90/72.35                  &  \ \ 4.62/3.45 &             \ \ 2.30/\textbf{\textcolor{blue}{1.27}}           &      \ 7.85/4.62         & 51.87/45.05 \\
			Early Fusion &   \textbf{\textcolor{red}{86.70}}/\textbf{\textcolor{red}{83.72}}  &   \textbf{\textcolor{blue}{64.17}}/\textbf{\textcolor{red}{60.49}}  &   \ \ 9.55/7.88 &  \textbf{\textcolor{blue}{15.67}}/\textbf{\textcolor{red}{13.33}} & \textbf{\textcolor{blue}{67.79}}/\textbf{\textcolor{red}{62.29}} \\ \hline
			Who2com   &      74.60/69.30                  &     \ \ 4.70/3.95                   &      \ \ \textbf{\textcolor{blue}{1.80}}/1.30                  &    \ 7.58/7.25                    &      49.77/42.30 \\
			When2com    &    75.10/69.69                  &   \ \ 4.79/4.08                     &     \ \ \textbf{\textcolor{green}{1.83}}/\textbf{\textcolor{green}{1.28}}                   &     \ 7.58/7.25                   &     52.06/44.21       \\
			V2VNet     &     80.13/75.17               &      44.66/32.00                 &      \ \ \textbf{\textcolor{red}{1.33}}/\textbf{\textcolor{red}{0.97}}               &     10.70/6.83                  & 58.51/48.98 \\
			DiscoNet &    \textbf{\textcolor{blue}{86.64}}/\textbf{\textcolor{blue}{82.16}}                    &   \textbf{\textcolor{green}{58.58}}/\textbf{\textcolor{green}{46.47}}                     &     \ \ 3.02/1.91                   &    \textbf{\textcolor{blue}{12.86}}/9.32                     & \textbf{\textcolor{green}{65.89}}/\textbf{\textcolor{green}{56.74}} \\ 
			Where2comm & 81.96/77.24 & 44.97/30.90 & 11.45/8.45 & \textbf{\textcolor{red}{16.51}}/\textbf{\textcolor{blue}{11.61}} & 62.05/54.59 \\ \hline
			UMC(ours) &      \textbf{\textcolor{green}{84.67}}/\textbf{\textcolor{green}{80.68}}        &      \textbf{\textcolor{red}{67.01}}/\textbf{\textcolor{blue}{60.04}}                  &        \ \ 2.38/1.70                &      12.35/\textbf{\textcolor{green}{9.46}}                 & \textbf{\textcolor{red}{67.80}}/\textbf{\textcolor{blue}{60.01}} \\ \hline
	\end{tabular}}
\end{minipage}
\hspace{3mm}%
\begin{minipage}[b]{0.48\linewidth} 
	\centering
	\caption{Detection comparison on OPV2V dataset. Key: [\textbf{\textcolor{red}{Best}}, \textbf{\textcolor{blue}{Second Best}}, \textbf{\textcolor{green}{Third Best}}]}
	\label{per-lambda}
	\resizebox{0.68\textwidth}{!}{
		\begin{tabular}{clll}
			\hline
			Models      & \ $\mathrm{ARSV}_{50/70}\uparrow$ & $\ \mathrm{ARCV}_{50/70}\uparrow$ & $\ \mathrm{AP}_{50/70}\uparrow$ \\ \hline
			No Fusion    &   69.33/46.35   &   12.32/ 4.36   &  54.69/23.94   \\
			Early Fusion &   64.62/46.95   &   \textbf{\textcolor{blue}{45.00}}/\textbf{\textcolor{blue}{24.39}}   &  55.88/\textbf{\textcolor{red}{25.89}} \\ \hline
			Who2com      &   69.43/45.68   &   15.84/6.21    &  55.36/23.39\\
			When2com     &   69.27/45.54   &   15.90/6.22    &  55.13/23.27 \\
			V2VNet       &   \textbf{\textcolor{blue}{73.42}}/\textbf{\textcolor{red}{48.21}}   &   24.65/11.16   &  \textbf{\textcolor{blue}{59.03}}/\textbf{\textcolor{green}{24.71}}  \\
			DiscoNet     &   \textbf{\textcolor{green}{72.04}}/46.94   &   \textbf{\textcolor{green}{37.25}}/\textbf{\textcolor{green}{22.57}}   &  55.46/23.04  \\
			Where2comm   &   69.79/\textbf{\textcolor{green}{47.60}}   &  13.89/6.09     &  \textbf{\textcolor{green}{56.02}}/\textbf{\textcolor{blue}{25.38}}   \\\hline
			UMC(ours)    &   \textbf{\textcolor{red}{76.56}}/\textbf{\textcolor{blue}{47.68}}   &   \textbf{\textcolor{red}{47.82}}/\textbf{\textcolor{red}{25.06}}   &  \textbf{\textcolor{red}{61.90}}/24.50  \\\hline
	\end{tabular}}
\end{minipage}
\end{table*}

\begin{figure*}[t]
\centering
\includegraphics[width=0.85\linewidth]{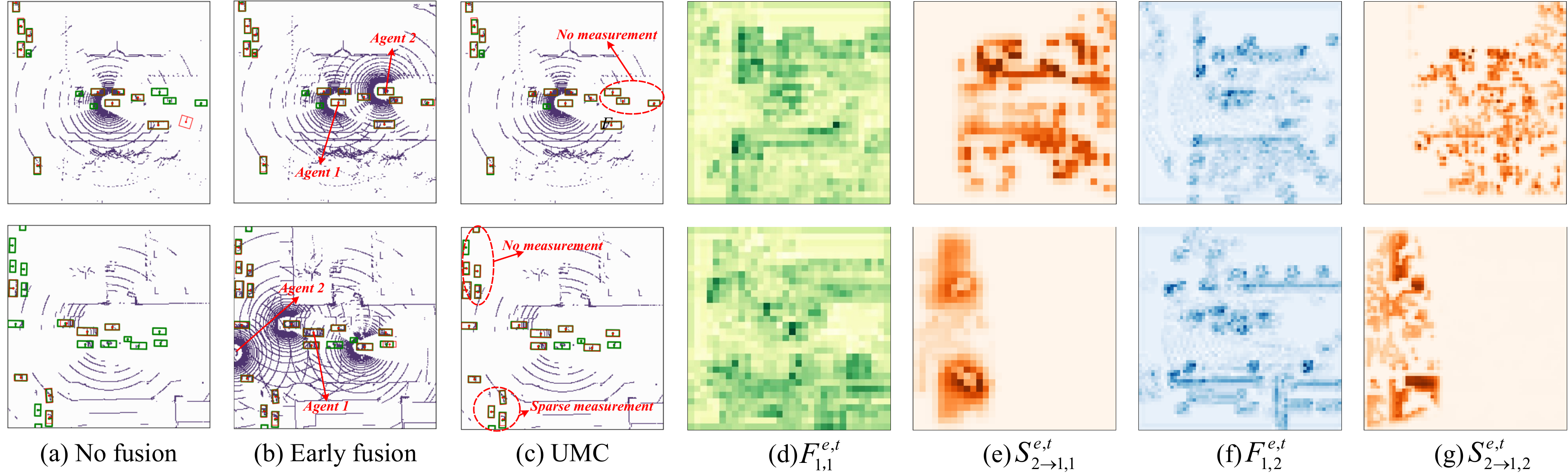}
\caption{Detection and communication selection for Agent 1. The green and red boxes represent the ground truth (GT) and predictions, respectively. (a-c) shows the results of no fusion, early fusion, and UMC compared to GT. (d) The coarse-grained collaborative feature of Agent 1. (e) Matrix-valued entropy-based selected communication coarse-grained feature map from Agent 2. (f) The fine-grained collaborative feature of Agent 1. (g) Matrix-valued entropy-based selected communication fine-grained feature map from Agent 2.}
\label{fig:big}
\end{figure*}

\subsection{Collaborative 3D Object detection dataset}

To validate our proposed UMC on the LIDAR-based 3D object detection task. Same as \cite{Li2021LearningDC, Lei2022LatencyAwareCP, li2022multi}, we utilize the multi-agent datasets of V2X-Sim\cite{li2022v2x} and OPV2V\cite{xu2022opencood}. The V2X-Sim is built by the co-simulation of SUMO\cite{SUMO2018} and CARLA\cite{Dosovitskiy17}, which contains 90 scenes, and each scene includes 100 frames. The OPV2V is co-simulated by OpenCDA\cite{xu2022opencood} and CARLA, including 12K frames of 3D point clouds with 230K annotated 3D boxes.


\subsection{Implementation details}
\noindent\textbf{Experimental setting} Same as \cite{Li2021LearningDC, Lei2022LatencyAwareCP, xu2022opencood}, we crop the point clouds located in the region of $[-32m,32m] \times [-32m,32m] \times [0,5m]$ defined in the ego-vehicle Cartesian coordinate system. The size of each voxel cell is set as $0.25m \times 0.25m \times 0.4m$ to get the Bird's-Eyes view map with dimension $256 \times 256 \times 13$. The SRAR-based transmitted collaborative feature map (TCF) has a dimension of $32 \times 32 \times 256$. Our proposed UMC's TCFs have the dimension of $32 \times 32 \times 256, 64 \times\ 64 \times 128$. We train all the models using NVIDIA RTX 3090 GPU with PyTorch\cite{paszke2017automatic}.

\noindent\textbf{Baselines} We consider the single-agent perception model, called \textit{No Fusion}, which only processes a single-view point cloud. Also, we consider the holistic-view perception model based on early collaboration, called \textit{Early Fusion}. We consider five collaboration methods: Who2com\cite{Liu2020Who2comCP}, When2com\cite{Liu2020When2comMP}, Where2comm\cite{Where2comm:22}, DiscoNet\cite{Li2021LearningDC} and V2VNet\cite{Wang2020V2VNetVC}. To ensure fairness, all the methods share the same encoder and detection header architecture and collaborate at the same intermediate feature layer.

\begin{figure}[t]
\centering
\includegraphics[width=0.85\linewidth]{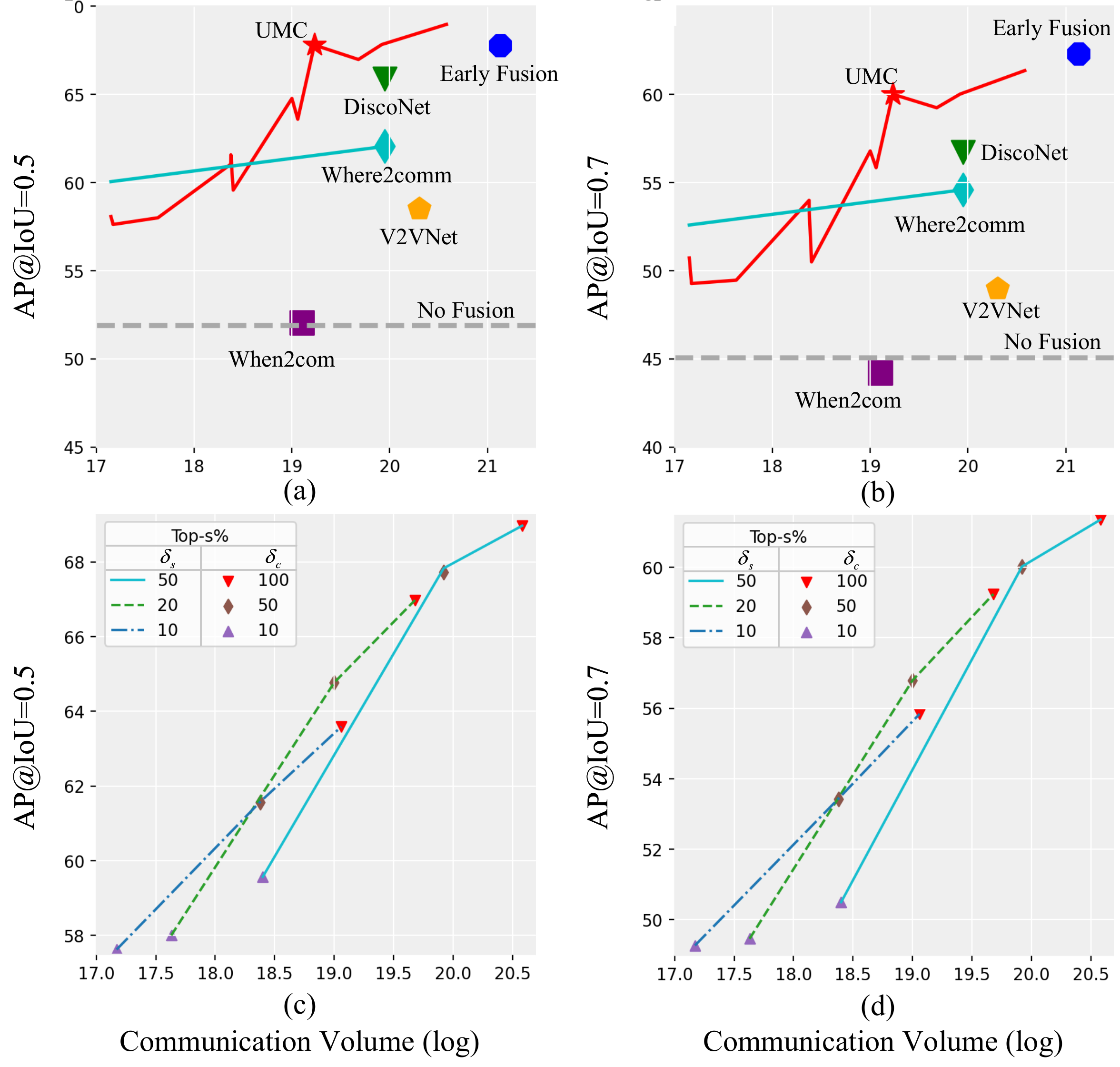}
\caption{Performance-bandwidth trade-off in log scale.} 
\label{fig:commu}
\end{figure}

\noindent\textbf{Evaluation Metrics} We employ the Average Precision (AP) to validate the overall performance of the model. 
To systematically analyze the performance of the model, we introduce new metrics from four aspects: i) \textbf{ARSV} denotes the \textbf{A}verage \textbf{R}ecall of agents that are visible from \textbf{S}inge-vehicle \textbf{V}iew; ii) \textbf{ARCV} denotes the \textbf{A}verage \textbf{R}ecall of agents that are invisible from single-vehicle view but visible from \textbf{C}ollaborative-\textbf{V}iew, which evaluates model's performance of collaborative process; iii) \textbf{ARCI} denotes the \textbf{A}verage \textbf{R}ecall of \textbf{C}ompletely-\textbf{I}nvisible agents, which evaluates the performance of detecting agents with limited information; iv) \textbf{ARTC} denotes the \textbf{A}verage \textbf{R}ecall of the agents that were visible at the last time but not visible at this time, which evaluates model's performance of \textbf{T}ime \textbf{C}ontinuity. Note that more technical details are shown in Appendix.1, and we employ all evaluation metrics at the Intersection-over-Union (IoU) threshold of 0.5 and 0.7. 

\subsection{Quantitative evaluation}

\noindent\textbf{Overall performance comparison} Table~\ref{table:v2x-sim} of $\mathrm{AP}_{50/70}$ column shows the comparisons in terms of AP(@IoU=0.5/0.7). We see that i) early fusion achieves the best detection performance when AP@0.7, and there is an obvious improvement over no fusion, \textit{i.e.}, AP@0.5 and AP@0.7 are increased by 30.69\% and 38.26\% respectively, revealing the effectiveness of collaboration; ii) among the SRAR-based intermediate collaboration methods, the proposed UMC achieves the best performance. Compared to When2com, UMC improves by 30.23\% in AP@0.5 and 35.71\% in AP@0.7. Compared to the knowledge distillation (KD) based DiscoNet, UMC improves by 2.9\% in AP@0.5 and 5.76\% in AP@0.7.

\begin{figure}[t]
\centering
\includegraphics[width=\linewidth]{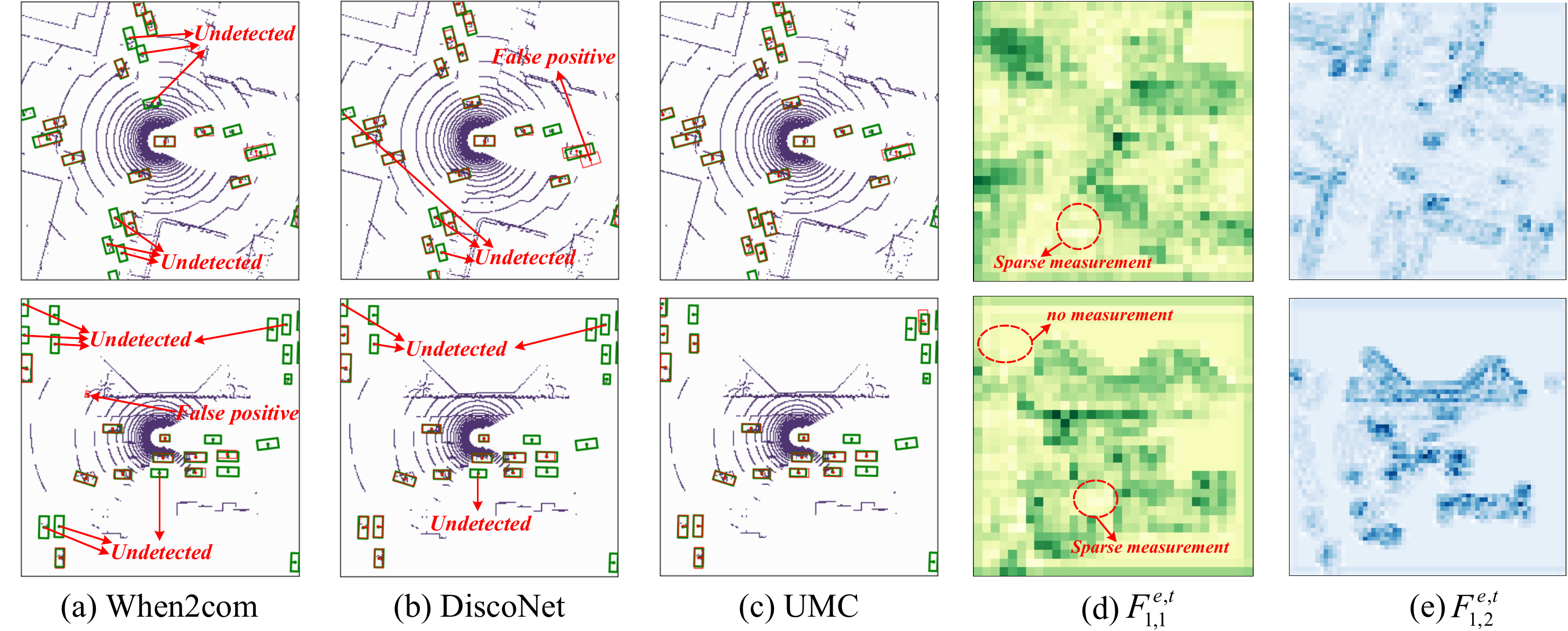}
\caption{UMC qualitatively outperforms the state-of-the-art methods. The green and red boxes denote ground truth and detection, respectively. (a) Results of When2com. (b) Results of DiscoNet. (c) Results of UMC. (d)-(e) Agent 1's coarse-grained and fine-grained collaborative feature maps, respectively.}
\label{fig:small}
\end{figure}

\noindent\textbf{Proposed new metric performance comparison} From the rest column of Tabel~\ref{table:v2x-sim}, we see that i) the ARSV reflects that collaboration can improve the performance of original single-view based objects. The early fusion achieves the best performance, and there is an obvious improvement over no fusion, \textit{i.e.}, ARSV@0.5 and ARSV@0.7 are increased by 12.74\% and 15.72\% respectively. Note that the DiscoNet inherits the performance of early fusion through knowledge distillation; ii) the ARCV reflects the quality of the collaboration process. The proposed UMC achieves the best performance. Compared to the DiscoNet, UMC improves by 14.39\% in ARCV@0.5 and 29.07\% in ARCV@0.7. Note that in when2com and who2com, the collaboration is simply integrated by stacking features along channels, resulting in the low quality of collaboration; iii) the ARCI reflects the detection performance under limited information. Thus, a high ARCI may cause more false positives, reducing the precision of the model; 
iv) the ARTC reflects the time continuity of the model, and the ARTC of existing methods is relatively low, In that sense, our work may unlock the future temporal reasoning in collaborative perception.


\noindent\textbf{Performance-bandwidth trade-off analysis} As demonstrated in Figure~\ref{fig:commu}, we comprehensively compare the proposed UMC under different ($\delta_{s},\delta_{c}$) values with the baseline methods in terms of the trade-off between performance and communication bandwidth. The no fusion model is shown as a dashed line since the communication volume is zero. We see that i) UMC achieves the best trade-off among the other methods; ii) the detect performance of where2comm is more stable than UMC when bandwidth is reduced. 
The where2comm is specific for the detection downstream task, which is based on the sparsity of foreground information to reduce communication by filtering the background intensively with a small performance sacrifice. And, our entropy-CS aims to optimize not only detection but also general downstream tasks based on the traditional information theory. Hence, when setting small value of ($\delta_{s},\delta_{c}$), the detection performance will degrade more easily than where2comm, shown in Figure~\ref{fig:commu} (c)-(d).

\subsection{Qualitative evaluation}

\begin{table}[t]
\centering
\caption{Detection performance with different grains selection. $\boldsymbol{F}^{e,t}_{i,n}(n=1,2,3)$ denotes different grained collaborative features. Note that the C. V. is the short for Communication Volume.}
\resizebox{0.35\textwidth}{!}{
\begin{tabular}{l|l|l}
\hline
\multicolumn{1}{c|}{\begin{tabular}[c]{@{}c@{}}\textbf{Multi-Grains Selection}\\ $\boldsymbol{F}^{e,t}_{i,1}$   \quad      $\boldsymbol{F}^{e,t}_{i,2}$  \quad  $\boldsymbol{F}^{e,t}_{i,3}$\end{tabular}} & $\mathrm{AP}_{50/70}\uparrow$ & C. V. $\downarrow$ \\ \hline
\quad  \checkmark  \quad  \quad \ \ \checkmark          &    \textbf{67.80}/\textbf{60.01}                  &         \textbf{19.23}                        \\
\quad  \checkmark  \quad  \quad \quad \  \quad \quad  \  \checkmark      &  58.36/50.71                       &         19.84                        \\ 
\quad \quad \quad  \quad \ \checkmark \quad \quad \  \checkmark &  60.27/53.63  &  20.02 \\ \hline
\end{tabular}
}
\label{table:ablation_2}
\end{table}

\begin{table*}[t]
\centering
\caption{Quantitative results of ablation experiments on V2X-Sim.}
\resizebox{0.85\textwidth}{!}{
\begin{tabular}{c|lll|lllll}
\hline
Variants & Entropy-CS & G-CGRU & MGFE & $\ \mathrm{ARSV}_{50/70}\uparrow$ & $\mathrm{ARCV}_{50/70}\uparrow$ & $\ \mathrm{ARCI}_{50/70}\downarrow$ &  $\mathrm{ARTC}_{50/70}\uparrow$ & $\ \ \mathrm{AP}_{50/70} \uparrow$ \\ \hline
(1)      &  \quad\quad\checkmark    &  \quad \  \ \checkmark  &   \quad \checkmark   &\ 86.67/80.68 & 67.01/60.04 & \ \ 2.38/1.70  &  12.35/9.46 & \textbf{67.80}/\textbf{60.01}   \\
(2)      &      & \quad \  \  \checkmark  &  \quad \checkmark   &\  87.67/83.29 &  67.88/62.17  & \ \ 2.87/1.66 & 12.01/9.55 &    65.72/58.78      \\
(3)      &   \quad\quad\checkmark   &   \quad \  \ \checkmark       &   &\   79.38/74.90 & 40.35/34.14  &  \ \ 1.49/1.28 & \ \ 3.79/3.52 &   56.73/49.22  \\ 
(4)      &   &    \multicolumn{1}{c}{\checkmark}        &                           &\ 87.72/82.13 & 60.58/46.53 &  \ \ 3.13/1.43  &  12.78/9.98 & 67.30/56.19                  \\ \hline
\end{tabular}
}
\label{table:ablation}
\end{table*}

\begin{figure*}[t]
\centering
\includegraphics[width=0.81\linewidth]{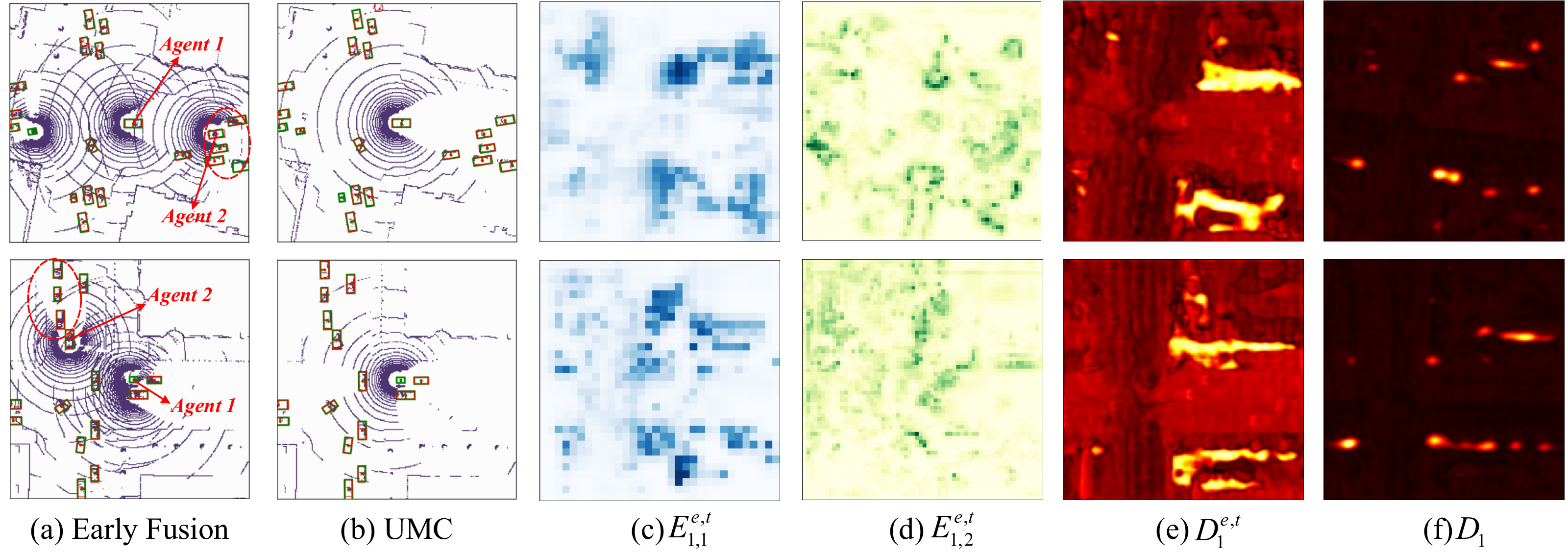}
\caption{Detection and feature map visualization for Agent 1. Green/red boxes represent GT/predictions. (a) and (b) denote the results of early fusion and UMC. (c) and (d) denote the coarse- and fine-grained collaborative maps generated from the G-CGRU module. respectively. (e) and (f) denote the UMC and early fusion's downstream features $\boldsymbol{D}_{1}$ for final detection output $\boldsymbol{\tilde{Y}_{1}}$, respectively.}
\label{fig:inter}
\end{figure*}

\noindent\textbf{Visualization of UMC working mechanism} To understand the working mechanism of the proposed UMC, we visualize the detection results, different grains collaborative feature maps, and the corresponding entropy-based transmission maps; see Figure~\ref{fig:big}. Note that the proposed entropy-based selection is matrix-based, reflecting the informative regions in a cell-level resolution, which is shown as the 2D communication map that is compressed along channels by a sum operation. Figure~\ref{fig:big} (a), (b), and (c) represent three detection results of Agent 1 based on the no fusion, early fusion, and the proposed UMC, respectively. We see that with collaboration, UMC is able to detect objects in these sheltered and long-range zones. To further explain the rationale behind the results, Figure~\ref{fig:big} (d) and (f) provide the corresponding ego coarse- and fine-grained collaborative feature maps, respectively. We clearly observe that the coarse-grained feature reflects the global structure information and the fine-grained feature contains rich local texture details, the proposed UMC complements their advantages. Figure~\ref{fig:big} (e) and (g) show the coarse- and fine-grained communication maps from agent 2 to agent 1, respectively. We observe that with the grains being finer, the communication map becomes more sparse, meaning the entropy-based selection can more precisely find the informative regions relative to ego agents, also, the entropy-based selection ensures lower bandwidth under the multi-grain collaborative process.

\noindent\textbf{Comparison with DiscoNet and When2com} Figure~\ref{fig:small} shows two examples to compare the proposed UMC with When2com and DiscoNet. We see that UMC is able to detect more objects, especially in edge regions with sparse measurement. The reason is that both When2com and DiscoNet employ only a single coarse-grained global feature map for collaborations, which loses many details of the scene, while UMC combines coarse- and fine-grained features for collaboration, see the visualization in Figure~\ref{fig:small} (d)-(e). Remarkably, compared to knowledge-distillation-based DiscoNet, the UMC does not need to pretrain different teacher models (early fusion) when switching to different datasets, and the performance will be influenced greatly when the teacher model is not performing well, as shown in Table~\ref{per-lambda}.

\subsection{Ablation study}

We validate several key designs of our framework, including the entropy-based communication selection, the G-CGRU, and the MGFE module. The quantitative results are shown in Table~\ref{table:ablation}. 

\noindent\textbf{Effect of grains selection} In tabel~\ref{table:ablation_2} we investigate the detection performance when employing different grain selections. Note that the shape of $\boldsymbol{F}^{e,t}_{i,n}(n=1,2,3) \in \mathbb{R}^{W,H,C}$ is $32 \times 32 \times 256$, $64 \times 64 \times 128$ and $128 \times 128 \times 64$, respectively. We see that i) applying collaborative feature maps on the selection of $(\boldsymbol{F}^{e,t}_{i,1},\boldsymbol{F}^{e,t}_{i,2})$ has the best trade-off between performance and bandwidth; and ii) the selection of $(\boldsymbol{F}^{e,t}_{i,1}, \boldsymbol{F}^{e,t}_{i,3})$ for collaboration has the worst detection performance. This is because the grains of feature difference between the two is too large, resulting in deviation in feature fusion process; and iii) the performance of selection of  $(\boldsymbol{F}^{e,t}_{i,2}, \boldsymbol{F}^{e,t}_{i,3})$ is slightly worse than $(\boldsymbol{F}^{e,t}_{i,1},\boldsymbol{F}^{e,t}_{i,2})$, because both $(\boldsymbol{F}^{e,t}_{i,2}, \boldsymbol{F}^{e,t}_{i,3})$ focus on the local texture details, resulting in the loss of global structure information; and iv) on the premise of performance of $(\boldsymbol{F}^{e,t}_{i,1},\boldsymbol{F}^{e,t}_{i,2})$, the heavy bandwidth of the $(\boldsymbol{F}^{e,t}_{i,1},\boldsymbol{F}^{e,t}_{i,2}, \boldsymbol{F}^{e,t}_{i,3})$ selection makes the potential performance improvement meaningless.

\noindent\textbf{Effect of collaboration strategy} Table~\ref{table:ablation} compares the different ablated networks. We see that i) the UMC (variants 1) achieves the best detection performance among the other ablated networks; and ii) variant 4 reflects that the redesigned graph-based collaborative GRU module has an obvious improvement over no fusion, \textit{i,e}, AP@0.5 and AP@0.7 are increased by 26.66\% and 24.72\% respectively, see the Figure~\ref{fig:inter} (c) and (d); and iii) variant 2 is composed of variants 4 and multi-grain feature enhance module. Compared to variant 4, variant 2 improves by 12.05\% in ARCV@0.5 and 33.61\% in ARCV@0.7, showing the effectiveness of the MGFE module. Meanwhile, as demonstrated in Figure~\ref{fig:inter} (e) and (f), the MGFE module can enhance the downstream feature map compared to the early fusion method; and iv) from the results of variants 3 and 4, and the results of variants 1 and 2, we observe that the MGFE module can give positive guidance to the trainable entropy-based communication selection module because the MGFE module allows the entropy-CS to comprehensively select regions from local and global viewpoints.

\section{Conclusion}
We propose a unified, bandwidth-efficient collaborative perception framework named UMC. Its core concept is to utilize the multi-resolution technique to cover the intrinsic interactions among the communication, collaboration, and reconstruction processes in MCP.
To validate it, we conduct experiments on the V2X-Sim and OPV2V datasets and propose a brand new evaluation metric. Comprehensive quantitative and qualitative experiments show that the UMC achieves an outstanding performance-bandwidth trade-off among the existing collaborative perception methods.


{\small
\bibliographystyle{ieee_fullname}
\bibliography{egbib}
}

\clearpage

\appendix

\section{Detailed information of Proposed Metrics}
\label{metrics}

\begin{figure}[t]
	\centering
	\includegraphics[width=\linewidth]{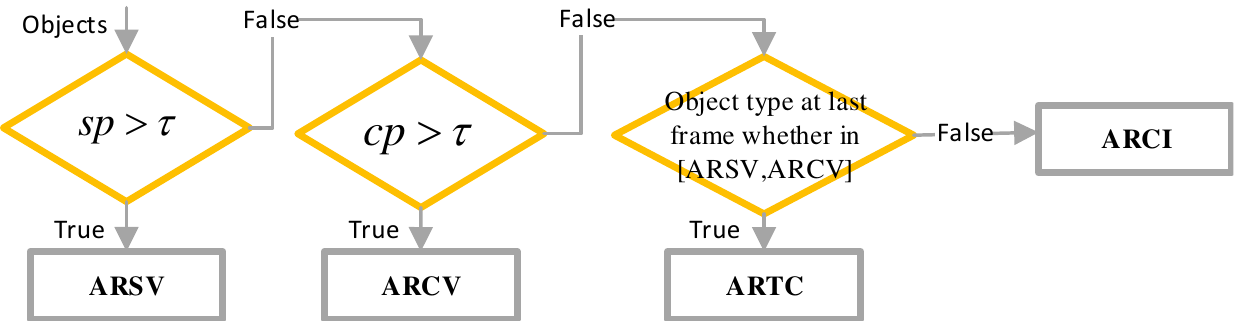}
	\caption{In the test set, we traverse all objects in each frame to obtain their corresponding types.\ Note that $sp, cp$ are short for detected points from single and collaborative view, respectively.}
	\label{fig:chartflow}
\end{figure}

\begin{figure}[t]
	\centering
	\includegraphics[width=\linewidth]{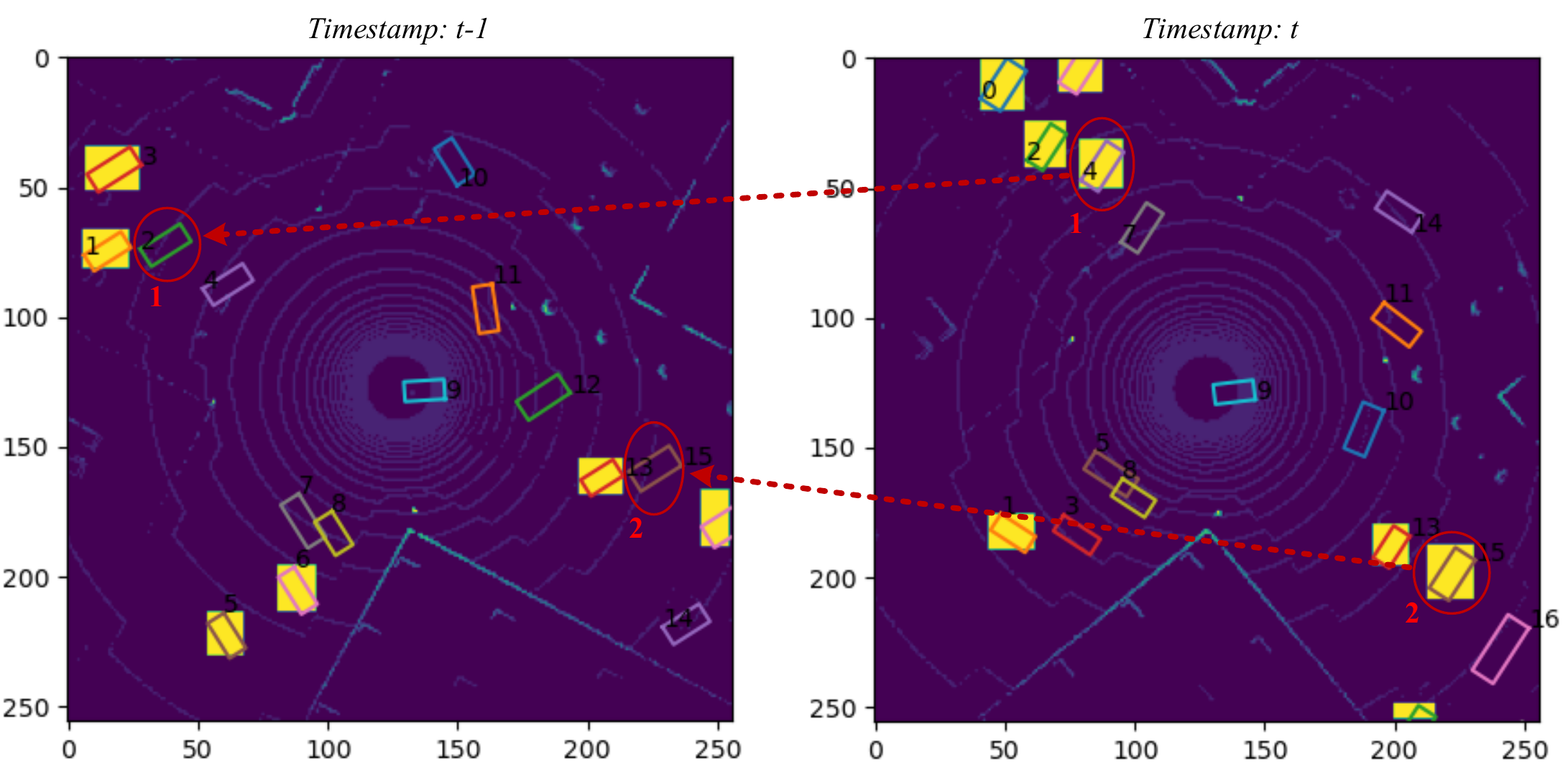}
	\caption{The detailed process of manually labelling.}
	\label{fig:details}
\end{figure}

The detailed process of proposed metrics are shown in Figure~\ref{fig:chartflow}. We traverse all objects in each frame to obtain their corresponding types. Note that the type of ARSV/ARCV can be automatically generated with code, as shown in Listing~\ref{code1}. And the type of ARCI/ARTC needs to be manually labeled, because whether it is visible at the last moment cannot be directly determined due to the movement of the vehicles from $t-1$ to $t$, as shown in Figure~\ref{fig:details}. 

For detailed process of manually labelling, \textit{e.g.}, at timestamp $t$, we firstly label vehicles with yellow background as ARCI type. Then, by comparing with time $t-1$, we find that agent \textcolor{red}{1,2} are visible at time $t-1$ (with no yellow background). Based on that, we change the type of agent \textcolor{red}{1,2} from ARCI to ARTC. The pseudo code are as follows:
\begin{lstlisting}[language=Python, caption=Pseudo code for proposed metrics, label=code1]
	single_view = np.load(...)
	collab_view = np.load(...)
	for object in len(single_view.objects):
	if singel_view[object].points > threhold:
	object.type = ARSV
	else:
	if collab_view[object].points > threhold:
	object.type = ARCV
	else:
	object.type = ARCI
	
	% Finally, we will manually label partial ARCI to ARTC, as shown in Figure 2. 
\end{lstlisting}

\begin{table}[t]
	\centering
	\caption{The ARSV and ARCV performance with different threshold points $\mathcal{\tau}$.}
	\resizebox{0.3\textwidth}{!}{
		\begin{tabular}{clllll}
			\hline
			Points     & $\mathrm{ARSV}_{50/70}$ & $\mathrm{ARCV}_{50/70}$   \\ \hline
			10 & 81.48/77.48  & 23.72/19.34 \\
			7 & 79.44/75.05 & 15.34/11.88 \\
			5 & 77.73/73.21 & \ \ 6.08/4.07 \\
			4  &      76.90/72.35                  &  \ \ 4.62/3.45  \\      \hline
		\end{tabular}
	}
	\label{table:v2x-sima}
\end{table}

As for the threshold points $\tau$, the proposed metrics distinguish between ARSV and ARCV based on whether they are visible. Nevertheless, whether they are visible depends on the number of points included in each object and the performance of the detector. To verify at how many points the detector can not detect the object, we conducted the following experiments on the \textit{No Fusion} model.

Table~\ref{table:v2x-sima} shows the ARSV and ARCV in terms of different points.  We can see that i) as the number of points decreases, so does the ARCV. This is because, as the points become more accurate, the no fusion model should theoretically be 0 in terms of ARCV; ii) when the points are greater than 5, the decline in ARCV is very large. When points are less than 5, the decline of the ARCV slows down, indicating it is close to the accurate points; iii) when points equal 4, the ARCV is 4.62\% in IoU@0.5 and 3.45\% in IoU@0.7, which are within the acceptable error range of 5\%. So, to decide if an object is visible, we look at how many points it has and whether that number is greater than 4. 

Note that the `last frame' of ARTC's time interval varies based on the sampling frequency (5Hz in V2X-Sim, 10Hz in OPV2V). And, the ARSV and ARCV only take recall into account, while the AP is weighted by both recall and precision. Hence, a high AP does not necessarily mean high ARSV or ARCV values. Based on that, you may wonder why not utilize APSV/APCV as the new metric. It is because the each proposed model can only predict the classifications (background or foreground) and regressions (x,y,w,h) of each pixel. Therefore, there is no prediction of the corresponding types for each detected objects.   

\section{Experiments details}

\subsection{Basic parameters}
\label{platform}

Our experiments are all performed on the workstation with AMD Core Ryzen Threadripper 3960X CPU and Nvidia 3090 GPU with Pytorch v1.7.1, CUDA 11.0. The SRAR-based’s transmitted
collaborative feature map (TCF) has a dimension of
32 × 32 × 256. Our proposed UMC’s TCFs have the dimension
of 32 × 32 × 256, 64 × 64 × 128. As for hyper-parameter tuning, we choose Adam as the optimizer and set the batch size to 4 for both V2X-Sim and OPV2V datasets. Also, we utilize the same number of scenes (total 80 scenes) for training. For testing stage, the V2X-sim utilizes 10 scenes, and OPV2V utilizes 15 scenes. Meanwhile, we use initial learning rate of 0.001 and set the random seed to 622.

\subsection{Baseline setting}

To ensure fairness, we fix the structure of the shared feature extractor MotionNet\cite{Wu2020jun} and detector, and transplant the collaborative part of the different methods without modification. Meanwhile, all the models are trained 100 epoch with initial learning rate of 0.001 and set the learning rate update strategy as `torch.optim.lr\_scheduler.MultiStepLR(self.
optimizer\_head, milestones=[50, 100], gamma=0.5)'.

\subsection{Communication Volume}

The communication volume for each method is calculated by:  $\mathrm{mean(log}(\sum^{agents}_{i=1}\sum^{T}_{t=1}(F.\mathrm{size}+Q.\mathrm{size})))$ during the test stage. $F$ represents the transmitted feature map, and $Q$ represents the query matrix, which is calculated in UMC, who2com, and when2com, and equals 0 in other methods.

\subsection{Setting of $\delta_{s},\delta_{c}$}

\begin{figure}[t]
	\centering
	\includegraphics[width=\linewidth]{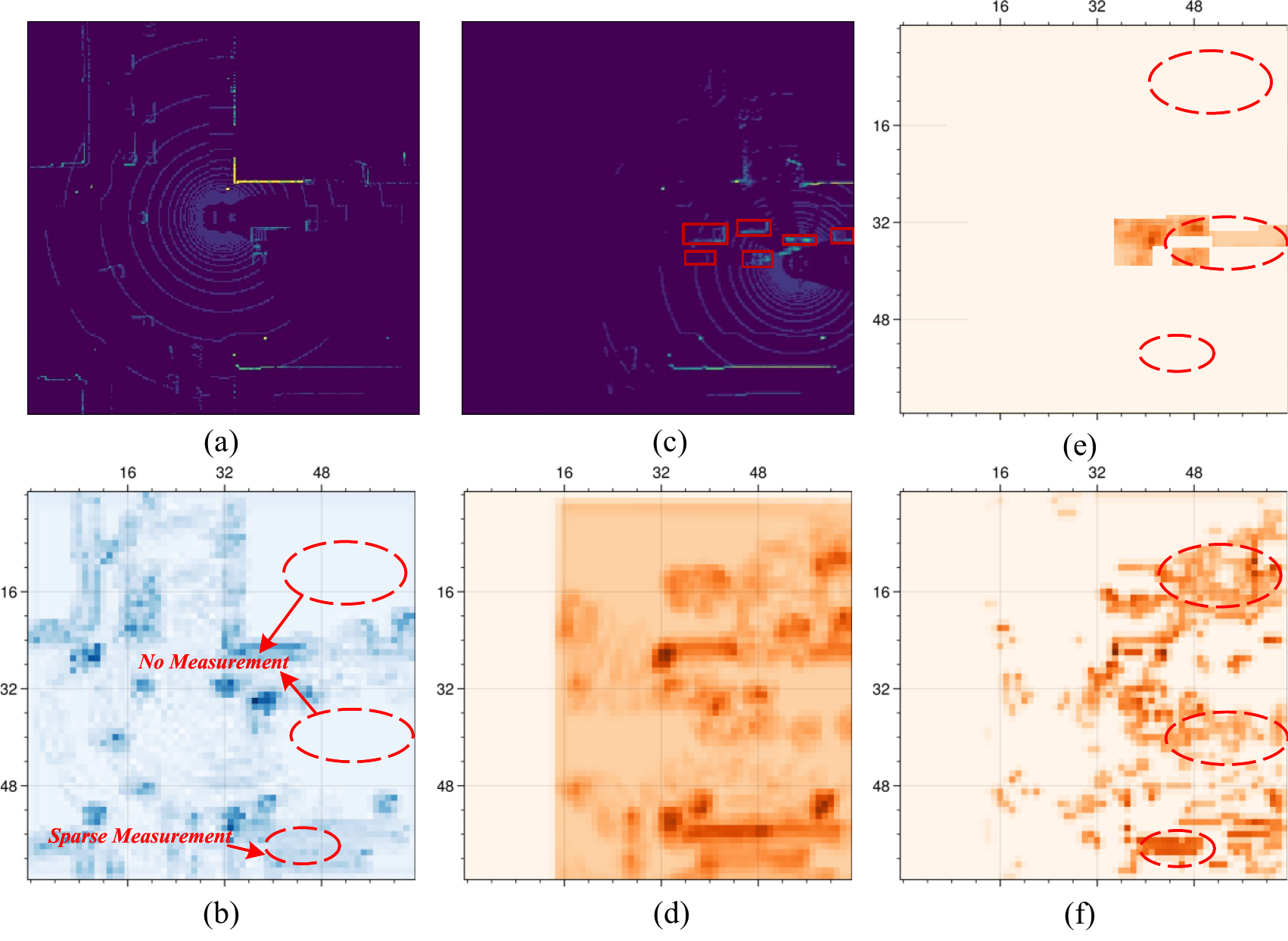}
	\caption{Difference between where2comm and UMC communication strategy. (a) The ego agent's observation.  (b) The ego agent's observation at feature level. (c) The collaborator's observation, red box denoted for the detected agents. (d) The collaborator's observation at feature level. (e) Communication map of where2comm. (f) Communication map of UMC.}
	\label{fig:diff}
\end{figure}

\begin{table}[h]
	\centering
	\caption{The performance-bandwidth trade off with different $\delta_{s},\delta_{c}$ values on V2X-Sim dataset.}
	\resizebox{0.45\textwidth}{!}{
        \begin{tabular}{lccccccc} 
        \hline
		\multicolumn{1}{c}{$\delta_{s}$} & $\delta_{c}$ & $\mathrm{ARSV}_{50/70}\uparrow$ & $\mathrm{ARCV}_{50/70} \uparrow$ & $\mathrm{ARCI}_{50/70}\downarrow$ &$\mathrm{ARTC}_{50/70}\uparrow$ & $\ \mathrm{AP}_{50/70} \uparrow $  & C.V. $\downarrow$  \\
  \hline
			50   & 100   & 85.66/81.87 & 70.76/63.15 & 2.41/1.7    & 11.88/8.12  & 68.97/61.35 & 20.58  \\
			50   & 50    & 84.78/80.73 & 67.46/60.72 & 2.50/1.78   & 11.88/8.12  & 67.83/60.02 & 19.92  \\
			50   & 10    & 80.62/75.39 & 31.48/23.53 & 2.17/1.37   & 12.03/7.20  & 59.57/50.50 & 18.4   \\
			20   & 100   & 87.30/84.08 & 74.08/66.77 & 16.68/12.24 & 22.15/17.41 & 66.98/59.24 & 19.68  \\
			20   & 50    & 83.23/78.98 & 56.49/48.87 & 2.12/1.51   & 11.23/8.08  & 64.77/56.79 & 19     \\
			20   & 10    & 80.53/75.38 & 19.02/12.33 & 2.14/1.38   & 9.16/6.74   & 58.00/49.46 & 17.63  \\
			10   & 100   & 86.57/82.97 & 59.37/51.68 & 14.57/10.51 & 19.09/15.15 & 63.59/55.84 & 19.06  \\
			10   & 50    & 82.12/77.44 & 38.52/31.28 & 2.16/1.47   & 10.77/8.35  & 61.57/53.43 & 18.376 \\
			10   & 10    & 80.78/75.51 & 13.29/8.97  & 2.21/1.39   & 10.77/8.35  & 57.62/49.27 & 17.171 \\
			5    & 100   & 86.04/82.59 & 47.19/40.44 & 14.17/9.9   & 19.25/15.14 & 61.04/53.99 & 18.375 \\
			5    & 20    & 80.87/75.72 & 15.09/10.76 & 2.12/1.36   & 10.77/8.35  & 57.67/49.45 & 17.17  \\
			1    & 100   & 85.44/81.49 & 34.59/28.72 & 13.47/9.88  & 20.49/17.35 & 58.07/50.72 & 17.15 \\
   \hline
		\end{tabular}
		\label{table:1}}
\end{table}

We record the proposed UMC under different ($\delta_{s},\delta_{c}$) in terms of the trade-off between performance and communication bandwidth, as shown in Table~\ref{table:1}.

Meanwhile, we comprehensively analyze the communication strategy between where2comm\cite{Where2comm:22} and our proposed UMC. As shown in Figure~\ref{fig:diff}, where2comm takes the advantages of sparsity of foreground information and only transmits the regions that the agents have. However, as for different downstream tasks, such as segmentation\cite{xu2022cobevt} or scene completion\cite{li2022multi}, there are other no-measurement or sparse-measurement regions that need collaborator's communication, as shown in Figure~\ref{fig:diff}.(b). Hence, our proposed UMC aim to optimize not only detection but also general downstream tasks based on the traditional information theory. From the Figure~\ref{fig:diff}.(f), we can observe that UMC can transmit the necessary regions to ego agent for general downstream tasks.

Note that in addition to discussing the top-$\delta$\% based filtering strategy of Eq.1 in manuscript, we also explored the mean based filtering strategy, the corresponding performance is shown as follows:

\begin{table}[h]
	\centering
	\caption{The performance of mean based filtering strategy on V2X-Sim dataset.}
	\resizebox{0.45\textwidth}{!}{
\begin{tabular}{lccccccc}
\hline
			\multicolumn{1}{c}{$\delta_{s}$} & $\delta_{c}$ & $\mathrm{ARSV}_{50/70}\uparrow$ & $\mathrm{ARCV}_{50/70} \uparrow$ & $\mathrm{ARCI}_{50/70}\downarrow$ &$\mathrm{ARTC}_{50/70}\uparrow$ & $\ \mathrm{AP}_{50/70} \uparrow $  & C.V. $\downarrow$      \\
   \hline
			mean &  & 84.67/80.68 & 67.01/60.04 & 2.38/1.70 & 12.35/9.46 & 67.80/60.01 & 19.23 \\
   \hline
	\end{tabular}}
\end{table}

\section{Performance analysis}

\subsection{Computation complexity}

The configuration of the experiment platform has been described in Section~\ref{platform}. Based on that, in terms of computations, the proposed entropy-cs only requires about 0.136G FLOPS with 0.66 ms latency to process a $256 \times\ 32 \times 32$ (C,H,W) feature map (more architecture details are shown in Section~\ref{entropy-cs}).

Compared to DiscoNet\cite{Li2021LearningDC}, the proposed C-GRU costs about 4.10G FLOPS more with 12.7 ms latency.

\subsection{Is Early Fusion always be better?}

We discuss the performance of early fusion model. As we all know, Early fusion aggregates the raw measurements from all collaborators, promoting a holistic
perspective. From the Table 1 in manuscript, the early fusion performs extremely good, even better than all the other baselines in some metrics. However, V2VNet\cite{Wang2020V2VNetVC} actually shows early fusion is far from optimal due to noises in real sensors. To address the above issue, since the dataset of V2VNet is not open source, we conduct experiments on OPV2V\cite{xu2022opencood} with Gaussian noises. As shown in Table~\ref{table:2}, we agree that the performance of Early Fusion may be degraded by noises to some extent.

\begin{table}[h]
	\centering
	\caption{Comparisons on OPV2V dataset.[\textbf{\textcolor{red}{Best}}, \textbf{\textcolor{blue}{Worst}}]}
	\resizebox{0.45\textwidth}{!}{
            \begin{tabular}{lccc}
            \hline
		\multicolumn{1}{c}{Method}       &    $\mathrm{ARSV}_{50/70}$         &     $\mathrm{ARCV}_{50/70}$        &     $\mathrm{AP}_{50/70}$        \\
  \hline
			No Fusion    & 69.33/\textbf{\textcolor{blue}{46.35}} & \textbf{\textcolor{blue}{12.32}}/\textbf{\textcolor{blue}{4.36}}  & \textbf{\textcolor{blue}{54.69}}/\textbf{\textcolor{blue}{23.94}} \\
			Early Fusion & \textbf{\textcolor{blue}{64.62}}/46.95 & 45.00/24.39 & 55.88/\textbf{\textcolor{red}{25.89}} \\
   \hline
			\multicolumn{1}{c}{UMC}          & \textbf{\textcolor{red}{76.56}}/\textbf{\textcolor{red}{47.68}} & \textbf{\textcolor{red}{47.82}}/\textbf{\textcolor{red}{25.06}} & \textbf{\textcolor{red}{61.90}}/24.50 \\\hline
	\end{tabular}}
	\label{table:2}
\end{table}

\subsection{Grains selection}

Table 3 in manuscript compares the performance of different selections of grain level. We also include comparisons of single-grain, as shown in Table~\ref{table:ablation_3}. Note that the heavy memory burden of all resolution baseline is not applicable on our RTX 3090.

\begin{table}[h]
	\centering
	\caption{Comparisons of single-grain selection.}
	\label{s1}
	\resizebox{0.35\textwidth}{!}{
		\begin{tabular}{l|l|l}
			\hline
			\multicolumn{1}{c|}{\begin{tabular}[c]{@{}c@{}}\textbf{Multi-Grains Selection}\\ $\boldsymbol{F}^{e,t}_{i,1}$   \quad      $\boldsymbol{F}^{e,t}_{i,2}$  \quad  $\boldsymbol{F}^{e,t}_{i,3}$\end{tabular}} & $\mathrm{AP}_{50/70}\uparrow$ & C. V. $\downarrow$ \\ \hline
			\quad \quad  \checkmark  \quad  \quad \ \          &    56.73/49.22  &  \ 7.88                       \\
			\quad  \quad \quad  \quad \quad \ \checkmark  \quad \quad &    58.96/52.86 & \ 8.12                                 \\ 
			\quad \quad \quad \quad  \quad \ \quad \ \quad \quad \  \checkmark &  57.43/51.66  &  \ 8.36  \\ \hline
		\end{tabular}
	}
	\label{table:ablation_3}
\end{table}

\section{More details about ablations analysis}

Table 4 in manuscript shows that a tremendous drop when adding Entropy-CS in variant 3 and 4. From our perspectives, variant 3 and 4 are based on single-resolution, then variant 3 (w/ entropy-cs)costs about $\frac{1}{4}$ communication of variant 4. Based on \cite{Where2comm:22}, when the communication is too small, the collaborative detection performance will suffer, resulting in a tremendous drop in variant 3. However, variant 3 still achieves detection gain compared with No Fusion (improved by $9.40\%/9.25\% \uparrow$ in $\mathrm{AP}_{50/70}$, respectively).

Meanwhile, AP of variant 2 is worse than variant 4 with comparable ARSV and better ARCV, this is because The ARSV and ARCV only take recall into account, while the AP is weighted by both recall and precision. Therefore, in variants 2 and 4, a high AP does not necessarily mean high ARSV or ARCV values, more details about proposed metrics can be found in Section~\ref{metrics}.

\section{Unified framework design}

We summarize the main contributions of recent collaborative algorithms in Table~\ref{tb:2}, , where $\checkmark$ indicates that a unique module is designed and $-$ indicates that general operations are utilized. Our proposed UMC optimizes the communication, collaboration, and reconstruction process with multi-resolution technique. 

\begin{table}[h]
	\centering
	\caption{Contribution summary.}
	\resizebox{0.35\textwidth}{!}{
		\begin{tabular}{lccc} 
            \hline
		\multicolumn{1}{c}{Method}                 & Comm. & Collab. & Recons. \\
            \hline
			Who2com (ICRA 2020~\cite{Liu2020Who2comCP})    &  \checkmark     &    -     & -        \\
			When2com (CVPR 2020~\cite{Liu2020When2comMP})   &  \checkmark     &    -     & -        \\
			V2VNet (ECCV 2020~\cite{Wang2020V2VNetVC})     &  -     &     \checkmark    & -        \\
			DiscoNet (NIPS 2021~\cite{Li2021LearningDC})   &   -    &  \checkmark       & -        \\
			V2X-ViT (ECCV 2022~\cite{Xu2022V2XViTVC})    &    -   &    \checkmark     & -        \\
			Where2comm (NIPS 2022~\cite{Where2comm:22}) &  \checkmark     &   \checkmark      &    -     \\
            \hline
		\multicolumn{1}{c}{UMC (ours)}             &  \checkmark     &   \checkmark      &   \checkmark\\
  \hline
	\end{tabular}}
	\label{tb:2}
\end{table}

\section{Detailed architecture of the model}

Note that we will release the source code.

\subsection{Architecture of entropy-CS}
\label{entropy-cs}

\begin{lstlisting}[language=Python, caption=Entropy-CS code]
	def acc_entropy_selection(self, tg_agent, nb_agent, delta1, delta2, M=3, N=3):
	self.stack = stack_channel(1, 9, kernel_size=3, padding=1)
	w = nb_agent.shape[-2]
	h = nb_agent.shape[-1]
	batch_nb = nb_agent.reshape(-1, 1, 1, 1) 
	stack = self.stack(nb_agent).permute(2,3,1,0).contiguous().reshape(-1, 9, 1, 1)
	
	p = F.sigmoid((stack - batch_nb)).mean(dim=1).reshape(w, h)
	entropy_tmp = p * torch.log(p)
	
	with torch.no_grad():
	top_delta = torch.sort(entropy_tmp.reshape(-1), descending=True)
	self_holder = top_delta[0][int(w*h*delta1)]
	
	masker = torch.where(entropy_tmp>=self_holder)
	
	stack_tg = self.stack(tg_agent).permute(2,3,1,0).contiguous().reshape(-1, 9, 1, 1)
	p_t = F.sigmoid((stack_tg - batch_nb)).mean(dim=1).reshape(w, h)
	entropy_t = p_t * torch.log(p_t)
	
	tmp_masker = - torch.ones_like(entropy_t)
	tmp_masker[masker] = entropy_t[masker]
	
	with torch.no_grad():
	top_delta2 = torch.sort(tmp_masker[tmp_masker!=-1].reshape(-1), descending=True)
	thresholds = top_delta2[0][int(w*h*delta2)]
	
	return torch.where(tmp_masker>=thresholds)
	
	
	class stack_channel(nn.Conv2d):
	def __init__(self, in_channels, out_channels, kernel_size, stride=1, padding=0, bias=False, interplate='none'):
	super(stack_channel, self).__init__(in_channels, out_channels, kernel_size=kernel_size, stride=stride, padding=padding, bias=bias)
	
	square_dis = np.zeros((out_channels, kernel_size, kernel_size))
	
	for i in range(out_channels):
	square_dis[i, i//3, i%3] = 1
	
	self.square_dis = nn.Parameter(torch.Tensor(square_dis), requires_grad=False)
	
	def forward(self, x):
	
	kernel = self.square_dis.detach().unsqueeze(1)
	stack = F.conv2d(x, kernel, stride=1, padding=1, groups=1)
	
	return stack
\end{lstlisting}

Our contribution of entropy-based selection is both theoretical and practical. The intuition of entropy-cs is low computational complexity and high interpretability. The entropy-cs is no parameter and single-round communication to reduce heavy bandwidth burden brought by multi-resolution technique.

\subsection{Architecture of G-CGRU}

To facilitate understanding, we have simplified many formulas and steps in manuscript. Hence, we add more technique details about the section of \verb|Graph-based Collaborative GRU|.

\begin{figure}[t]
	\centering
	\includegraphics[width=0.7\linewidth]{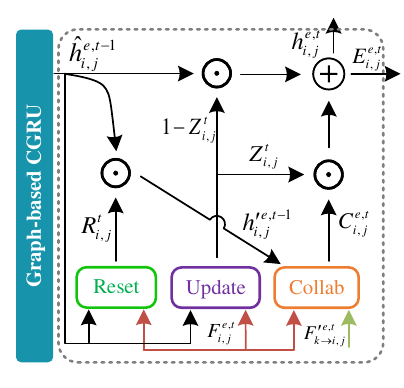}
	\caption{The architecture of G-CGRU.}
	\label{fig:ggrcu}
\end{figure}

For the ego agent $i$ of the $j$-th resolution intermediate feature maps, the inputs of G-CGRU are hidden states $\boldsymbol{h}^{e,t-1}_{i,j}$, the ego agent observation $\boldsymbol{F}^{e,t}_{i,j}$, and the supporters' selected feature maps $\{\boldsymbol{F}^{'e,t}_{k \to i,j}\}_{k \ne i}$, then the updates for G-CGRU at $t$-th step can be formulated as:
\abovedisplayshortskip=0pt
\belowdisplayshortskip=0pt
\begin{equation}
	\begin{aligned}
		& \boldsymbol{\hat{h}}^{e,t-1}_{i,j} = \Lambda(\boldsymbol{h}^{e,t-1}_{i,j}, \boldsymbol{\xi}^{t-1 \to t}_{i}) \\
		& \boldsymbol{R}^{t}_{i,j} = Reset(\boldsymbol{F}^{e,t}_{i,j}, \boldsymbol{\hat{h}}^{e,t-1}_{i,j})  \\ 
		& \boldsymbol{Z}^{t}_{i,j} = Update(\boldsymbol{\hat{h}}^{e,t-1}_{i,j},\boldsymbol{F}^{e,t}_{i,j}) \\ 
		& \boldsymbol{h}^{'e,t-1}_{i,j} = 
		\boldsymbol{\hat{h}}^{e,t-1}_{i,j}  \odot \boldsymbol{R}^{t}_{i,j}\\
		& \boldsymbol{C}^{e,t}_{i,j} = Collab(\boldsymbol{h}^{'e,t-1}_{i,j},\{\boldsymbol{F}^{'e,t}_{k \to i,j}\}_{k \ne i}, \boldsymbol{F}^{e,t}_{i,j}) \\
		& \boldsymbol{E}^{e,t}_{i,j} = \boldsymbol{Z}^{t}_{i,j} \odot \boldsymbol{C}^{e,t}_{i,j} + (1 -  \boldsymbol{Z}^{t}_{i,j}) \odot \boldsymbol{\hat{h}}^{e,t-1}_{i,j} \\
		& \boldsymbol{h}^{e,t}_{i,j} = \boldsymbol{W}_{3 \times 3}*\boldsymbol{E}^{e,t}_{i,j}
	\end{aligned}
	\label{eq:collob}
\end{equation}

where $\odot$, $*$ represent dot product and $3\times3$ convolution operation, respectively. $\boldsymbol{W}_{3 \times 3}$ indicates trainable parameters. The last time hidden feature $\boldsymbol{h}^{e,t-1}_{i,j}$ needs conduct feature alignment operation from $t-1$ to $t$ to get $\boldsymbol{\hat{h}}^{e,t-1}_{i,j}$, which ensure the $\boldsymbol{\hat{h}}^{e,t-1}_{i,j}$ and $\boldsymbol{F}^{e,t}_{i,j}$ are supported in the same coordinate system.

The $Reset$ and $Update$ gate modules share the same structure. Here we take $Reset$ as an example: 

\abovedisplayshortskip=0pt
\belowdisplayshortskip=0pt
\begin{equation}
	\begin{aligned}
		& \boldsymbol{W}_{ir} = \sigma(\boldsymbol{W}_{3 \times 3}*([\boldsymbol{\hat{h}}^{e,t-1}_{i,j};\boldsymbol{F}^{e,t}_{i,j}])) \\
		& \boldsymbol{R}^{t}_{i,j} = \sigma(\boldsymbol{W}_{ir} \odot \boldsymbol{\hat{h}}^{e,t-1}_{i,j} + (1 - \boldsymbol{W}_{ir}) \odot \boldsymbol{F}^{e,t}_{i,j})
	\end{aligned}
	\label{eq:reset1}
\end{equation}

where $\sigma(\cdot), [\cdot;\cdot]$ represent Sigmoid function and concatenation operation along channel dimensions. 
The gate $\boldsymbol{R}^{t}_{i,j} \in \mathbb{R}^{K,K,C}$ learns where the hidden features $\boldsymbol{h}^{e,t-1}_{i,j}$ are conducive to the present.

Based on the above $Reset$ and $Update$ modules, we thus derive the $Collab$ module. To make better collaborative feature integration, we construct a collaboration graph $\mathcal{G}^{t}_{c}(\boldsymbol{\mathcal{V}},\boldsymbol{\mathcal{E}})$ in $Collab$ module, where node $\boldsymbol{\mathcal{V}} = \{\mathcal{V}_{i}\}_{i=1,\dots,N}$ is the set of collaborative agents in environment and $\boldsymbol{\mathcal{E}} = \{\boldsymbol{W}_{i \to j}\}_{i,j=1,\dots,N}$ is the set of trainable edge matrix weights between agents and models the collaboration strength between two agents. Let $\mathcal{C}_{\mathcal{G}^{t}_{c}}(\cdot)$ be the collaboration process defined in the $Collab$ module's graph $\mathcal{G}^{t}_{c}$. The $j$-th resolution enhanced maps of ego $i$ agent after collaboration are $\boldsymbol{E}^{e,t}_{i,j} \gets \mathcal{C}_{\mathcal{G}^{t}_{c}}(\boldsymbol{h}^{e,t-1}_{i,j}, \boldsymbol{F}^{'e,t}_{k \to i,j}, \boldsymbol{F}^{e,t}_{i,j})$. This process has two stages: message attention (\textbf{S1}) and message aggregation (\textbf{S2}).

\begin{equation}
	\begin{aligned}
		\label{eq:collab}
		& \boldsymbol{W}_{k \to i} = \Pi([\boldsymbol{h}^{'e,t-1}_{i,j};\boldsymbol{F}^{'e,t}_{k \to i,j};\boldsymbol{F}^{e,t}_{i,j}] \in \mathbb{R}^{K,K} \\
		& \boldsymbol{\bar{W}}_{k \to i} = \frac{e^{\boldsymbol{W}_{k \to i}}}{\sum^{N}_{m=1}e^{\boldsymbol{W}^{t}_{m \to i}}} \\
		& \boldsymbol{C}^{e,t}_{i,j} = \sum^{N}_{m=1} \boldsymbol{\bar{W}}_{m \to i} \circ \boldsymbol{F}^{'e,t}_{m \to i,j}
	\end{aligned}
\end{equation}

In the \textit{message attention stage} (\textbf{S1}), each agent determines the matrix-valued edge weights, which reflect the strength from one agent to another at each individual cell. To determine the edge weights, we firstly get the conductive history information from hidden features by $Reset$ gates through $\boldsymbol{h}^{'e,t-1}_{i,j} \gets \boldsymbol{\hat{h}}^{e,t-1}_{i,j}  \odot \boldsymbol{R}^{t}_{i,j}$. Then, we utilize the edge encode $\Pi$ to correlate the history information, the feature map from another agent and ego feature map; that is, the matrix-value edge weight from $k$-th agent to the $i$-th agent is $\boldsymbol{W}_{k \to i}=\Pi(\boldsymbol{h}^{'e,t-1}_{i,j},\boldsymbol{F}^{'e,t}_{k \to i,j},\boldsymbol{F}^{e,t}_{i,j}) \in \mathbb{R}^{K,K}$, where $\Pi$ concatenates three feature maps along the channel dimension and then utilizes four $1 \times 1$ convolutional layers to gradually reduce the number of channels from $3C$ to 1, more details are shown in Section~\ref{edge_encoder}. Also, to normalize the edge weights across different agents, we implement a softmax operation on each cell of the feature map.

In the \textit{message aggregation stage} (\textbf{S2}), each agent aggregates the feature maps from collaborators based on the normalized matrix-valued edge weights, the updated feature map $\boldsymbol{C}^{e,t}_{i,j}$ is utilized by $\sum^{N}_{k=1} \boldsymbol{W}_{k \to i} \circ \boldsymbol{F}^{'e,t}_{k \to i,j}$, where $\circ$ represents the dot production broadcasting along the channel dimension. 

Finally, the collaborative map is $\boldsymbol{E}^{e,t}_{i,j} = \boldsymbol{Z}^{t}_{i,j} \odot \boldsymbol{C}^{e,t}_{i,j} + (1 -  \boldsymbol{Z}^{t}_{i,j}) \odot \boldsymbol{h}^{e,t-1}_{i,j}$. Note that the $\boldsymbol{Z}^{t}_{i,j}$ is generated by $Update$ gate and $\odot$ is the dot product. And, the hidden state is updated as $\boldsymbol{h}^{e,t}_{i,j} \gets \boldsymbol{W}_{3 \times 3}*\boldsymbol{E}^{e,t}_{i,j}$.


\subsection{Architecture of shared encoder}


We use the main architecture of MotionNet\cite{Wu2020jun} as our shared encoder. The input BEV map's dimension is $(c,w,h)=(13,256,256)$. We describe the architecture of the encoder below:

\begin{lstlisting}[language=Python, caption=Shared encoder code]
	nn.Sequential(
	nn.Conv2d(13, 32, 3, stride=1, padding=1)
	nn.BatchNorm2d(32)
	nn.ReLU()
	nn.Conv2d(32, 32, 3, stride=1, padding=1)
	nn.BatchNorm2d(32)
	nn.ReLU()
	nn.Conv3D(64, 64, (1,1,1), stride=1)
	nn.Conv3D(128, 128, (1,1,1), stride=1)
	nn.Conv2d(32, 64, 3, stride=2, padding=1)
	nn.BatchNorm(64)
	nn.ReLU()
	nn.Conv2d(64, 128, 3, stride=2, padding=1)
	nn.BatchNorm(128)
	nn.ReLU()
	nn.Conv2d(128, 128, 3, stride=1, padding=1)
	nn.BatchNorm(128)
	nn.ReLU()
	nn.Conv2d(128, 256, 3, stride=2, padding=1)
	nn.BatchNorm(256)
	nn.ReLU()
	nn.Conv2d(256, 256, 3, stride=1, padding=1)
	nn.BatchNorm(256)
	nn.ReLU()
	nn.Conv2d(256, 512, 3, stride=2, padding=1)
	nn.BatchNorm(512)
	nn.ReLU()
	nn.Conv2d(512, 512, 3, stride=1, padding=1)
	nn.BatchNorm(512)
	nn.ReLU())
\end{lstlisting}
\subsection{Architecture of SRAR-based shared decoder}

The input of the SRAR-based shared decoder is the intermediate feature output by each layer of the encoder. Its architecture is shown below:

\begin{lstlisting}
	nn.Sequential(
	nn.Conv2d(512 + 256, 256, 3, 1, 1)
	nn.BatchNorm2d(256)
	nn.ReLU()
	nn.Conv2d(256, 256, 3, 1, 1)
	nn.BatchNorm2d(256)
	nn.ReLU()
	nn.Conv2d(256 + 128, 128, 3, 1, 1)
	nn.BatchNorm2d(128)
	nn.ReLU()
	nn.Conv2d(128, 128, 3, 1, 1)
	nn.BatchNorm2d(128)
	nn.ReLU()
	nn.Conv2d(128 + 64, 64, 3, 1, 1)
	nn.BatchNorm2d(64)
	nn.ReLU()
	nn.Conv2d(64, 64, 3, 1, 1)
	nn.BatchNorm2d(64)
	nn.ReLU()
	nn.Conv2d(64 + 32, 32, 3, 1, 1)
	nn.BatchNorm2d(32)
	nn.ReLU()
	nn.Conv2d(32, 32, 3, 1, 1)
	nn.BatchNorm2d(32)
	nn.ReLU())
\end{lstlisting}

\subsection{Architecture of query generator}

The entropy-CS compresses the intermediate feature to generate query matrix by query generator, which is for light communication. Its architecture is shown below:

\begin{lstlisting}[language=Python, caption=Query generator code]
	nn.Sequential(
	nn.Conv2d(256, 64, 1, 1, 0)
	nn.BatchNorm2d(64)
	nn.ReLU()
	nn.Conv2d(64, 1, 1, 1, 0)
	nn.ReLU())
\end{lstlisting}

\subsection{Architecture of edge encoder $\Pi$}
\label{edge_encoder}

\begin{lstlisting}[language=Python, caption=Edge encoder code]
	class EdgeEncoder(nn.Module):
	def __init__(self,channel):
	super(EdgeEncoder, self).__init__()
	
	self.conv1_1 = nn.Conv2d(channel, 128, kernel_size=1, stride=1, padding=0)
	self.bn1_1 = nn.BatchNorm2d(128)
	
	self.conv1_2 = nn.Conv2d(128, 32, kernel_size=1, stride=1, padding=0)
	self.bn1_2 = nn.BatchNorm2d(32)
	
	self.conv1_3 = nn.Conv2d(32, 8, kernel_size=1, stride=1, padding=0)
	self.bn1_3 = nn.BatchNorm2d(8)
	
	self.conv1_4 = nn.Conv2d(8, 1, kernel_size=1, stride=1, padding=0)
	
	def forward(self, x):
	x = x.view(-1, x.size(-3), x.size(-2), x.size(-1))
	x_1 = F.relu(self.bn1_1(self.conv1_1(x)))
	x_1 = F.relu(self.bn1_2(self.conv1_2(x_1)))
	x_1 = F.relu(self.bn1_3(self.conv1_3(x_1)))
	x_1 = F.relu(self.conv1_4(x_1))
	
	return x_1
\end{lstlisting}

\begin{figure*}[!t]
	\centering
	\includegraphics[width=0.9\linewidth]{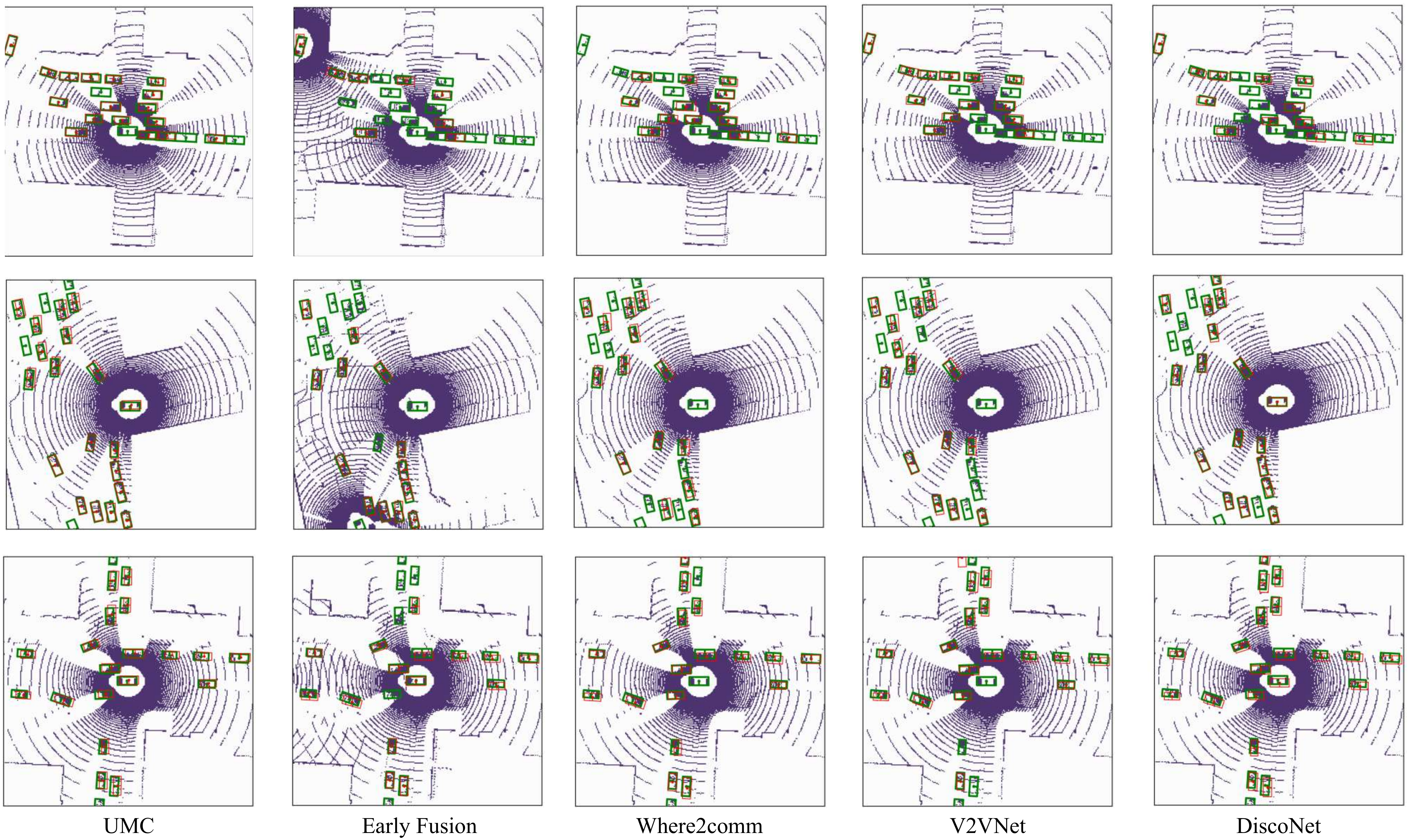}
	\caption{Detection results of UMC, Early Fusion, Where2comm, V2VNet and DiscoNet on OPV2V dataset.}
	\label{fig:big1}
\end{figure*}

\begin{figure*}[!t]
	\centering
	\includegraphics[width=0.9\linewidth]{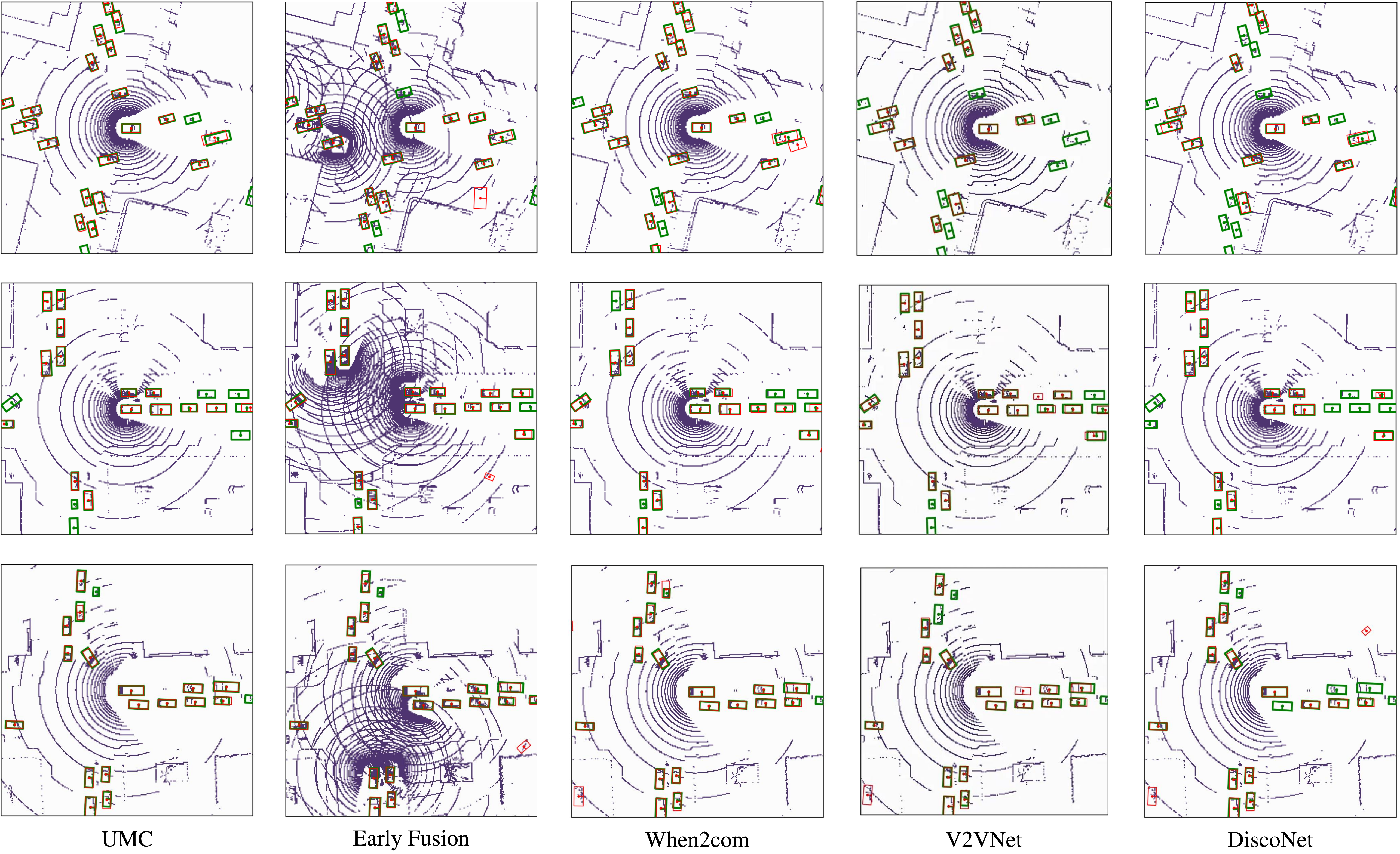}
	\caption{Detection results of UMC, Early Fusion, When2com\cite{Liu2020When2comMP}, V2VNet and DiscoNet on V2X-Sim dataset.}
	\label{fig:big2}
\end{figure*}

\subsection{Architecture of MGFE}

\begin{lstlisting}[language=Python, caption=MGFE code]
	class MGFE(nn.Module):
	def __init__(self, input_channel):
	super(MGFE, self).__init__()
	
	self.guide_v1 = nn.Conv2d(128, 128, kernel_size=1, stride=1, padding=0)
	self.guide_v1_bn = nn.BatchNorm2d(128)
	self.guide = nn.Conv2d(256, 256, kernel_size=1, stride=1, padding=0)
	self.guide_bn = nn.BatchNorm2d(256)
	
	self.conv5_1 = nn.Conv2d(512 + 256 + 256, 256, kernel_size=3, stride=1, padding=1)
	self.bn5_1 = nn.BatchNorm2d(256)
	
	self.conv5_2 = nn.Conv2d(256, 256, kernel_size=3, stride=1, padding=1)
	self.bn5_2 = nn.BatchNorm2d(256)
	
	self.conv6_1 = nn.Conv2d(256 + 128 + 128, 128, kernel_size=3, stride=1, padding=1)
	self.bn6_1 = nn.BatchNorm2d(128)
	
	self.conv6_2 = nn.Conv2d(128, 128, kernel_size=3, stride=1, padding=1)
	self.bn6_2 = nn.BatchNorm2d(128)
	
	self.conv7_1 = nn.Conv2d(128 + 64, 64, kernel_size=3, stride=1, padding=1)
	self.conv7_2 = nn.Conv2d(64, 64, kernel_size=3, stride=1, padding=1)
	
	self.conv8_1 = nn.Conv2d(64 + 32, 32, kernel_size=3, stride=1, padding=1)
	self.conv8_2 = nn.Conv2d(32, 32, kernel_size=3, stride=1, padding=1)
	
	self.bn7_1 = nn.BatchNorm2d(64)
	self.bn7_2 = nn.BatchNorm2d(64)
	
	self.bn8_1 = nn.BatchNorm2d(32)
	self.bn8_2 = nn.BatchNorm2d(32)
	
	self.norm1 = L2Norm(512)
	self.norm2 = L2Norm(256)
	self.norm3 = L2Norm(128)
	self.norm4 = L2Norm(64)
	self.norm5 = L2Norm(32)
	
	def forward(self, x, x_1, x_2, x_3, x_4, enhance_v1, enhance, batch, kd_flag = 0):
	
	enhance_v1 = enhance_v1.view(batch, -1, enhance_v1.size(1), enhance_v1.size(2), enhance_v1.size(3))
	enhance_v1 = enhance_v1.permute(0, 2, 1, 3, 4).contiguous()
	enhance_v1 = enhance_v1.permute(0, 2, 1, 3, 4).contiguous()
	enhance_v1 = enhance_v1.view(-1, enhance_v1.size(2), enhance_v1.size(3), enhance_v1.size(4)).contiguous()
	
	guide_v1 = torch.max(F.relu(self.guide_v1_bn(self.guide_v1(enhance_v1))), dim=1, keepdim=True)[0]
	guide = torch.max(F.relu(self.guide_bn(self.guide(enhance))), dim=1, keepdim=True)[0]
	x_3_guide = guide * x_3
	
	x_5 = F.relu(self.bn5_1(self.conv5_1(torch.cat((self.norm1(F.interpolate(x_4, scale_factor=(2, 2))), self.norm2(x_3_guide), self.norm2(enhance)), dim=1))))
	x_5 = F.relu(self.bn5_2(self.conv5_2(x_5)))
	
	x_2 = x_2.view(batch, -1, x_2.size(1), x_2.size(2), x_2.size(3))
	x_2 = x_2.permute(0, 2, 1, 3, 4).contiguous()
	x_2 = x_2.permute(0, 2, 1, 3, 4).contiguous()
	x_2 = x_2.view(-1, x_2.size(2), x_2.size(3), x_2.size(4)).contiguous()
	
	x_2_guide = guide_v1 * x_2
	
	x_6 = F.relu(self.bn6_1(self.conv6_1(torch.cat((self.norm2(F.interpolate(x_5, scale_factor=(2, 2))), self.norm3(x_2_guide), self.norm3(enhance_v1)), dim=1))))
	
	x_6 = F.relu(self.bn6_2(self.conv6_2(x_6)))
	
	x_1 = x_1.view(batch, -1, x_1.size(1), x_1.size(2), x_1.size(3))
	x_1 = x_1.permute(0, 2, 1, 3, 4).contiguous()
	x_1 = x_1.permute(0, 2, 1, 3, 4).contiguous()
	x_1 = x_1.view(-1, x_1.size(2), x_1.size(3), x_1.size(4)).contiguous()
	
	x_7 = F.relu(self.bn7_1(self.conv7_1(torch.cat((self.norm3(F.interpolate(x_6, scale_factor=(2, 2))), self.norm4(x_1)), dim=1))))
	x_7 = F.relu(self.bn7_2(self.conv7_2(x_7)))
	
	
	x = x.view(batch, -1, x.size(1), x.size(2), x.size(3))
	x = x.permute(0, 2, 1, 3, 4).contiguous()
	x = x.permute(0, 2, 1, 3, 4).contiguous()
	x = x.view(-1, x.size(2), x.size(3), x.size(4)).contiguous()
	
	x_8 = F.relu(self.bn8_1(self.conv8_1(torch.cat((self.norm4(F.interpolate(x_7, scale_factor=(2, 2))), self.norm5(x)), dim=1))))
	res_x = F.relu(self.bn8_2(self.conv8_2(x_8)))
	
	return res_x
	
	class L2Norm(nn.Module):
	def __init__(self, n_channels, scale=10.0):
	super(L2Norm, self).__init__()
	self.n_channels = n_channels
	self.scale = scale
	self.eps = 1e-10
	self.weight = nn.Parameter(torch.Tensor(self.n_channels))
	self.weight.data *= 0.0
	self.weight.data += self.scale
	
	def forward(self, x):
	norm = x.pow(2).sum(dim=1, keepdim=True).sqrt() + self.eps
	x = x / norm * self.weight.view(1, -1, 1, 1)
	
	return x
	
\end{lstlisting}

\section{Detailed information of Interpolation}

We utilize the main architecture of ADP-C\cite{liu2022anytime} as our interpolate function in entropy-CS module. We assume the input feature as $\boldsymbol{f}_{in} \in \mathbb{R}^{K,K,C}$ and suppose the pixels of the $\boldsymbol{f}_{in}$ are indexed by $\boldsymbol{p}$. Then, we form a mask $\boldsymbol{M}\in \mathbb{R}^{K,K}$:

\begin{equation}
	\boldsymbol{M}(\boldsymbol{p}) = 
	\begin{cases}
		0, & \text{if} \ \boldsymbol{f}_{in}(\boldsymbol{p}) = \boldsymbol{0}\\
		1, & \text{otherwise}
	\end{cases}  
\end{equation}

Assuming $C$ is a convolution layer with input $\boldsymbol{f}_{in}$, the by applying the mask, the output $\boldsymbol{f}_{out}$ at position $\boldsymbol{p}$ becomes:

\begin{equation}
	\boldsymbol{f}_{out}(\boldsymbol{p}) = 
	\begin{cases}
		C(\boldsymbol{f}_{in})(\boldsymbol{p}), & \text{if} \ \boldsymbol{M}(\boldsymbol{p})=1,\\
		\boldsymbol{0}, & \text{if}  \ \boldsymbol{M}(\boldsymbol{p})=0.
	\end{cases}  
\end{equation}

Denoting the interpolation operation as $\boldsymbol{I}$, the final output feature $\boldsymbol{f}^{*}_{out}$ is:

\begin{equation}
	\boldsymbol{f}^{*}_{out}(p) = 
	\begin{cases}
		\boldsymbol{f}_{out}(\boldsymbol{p}), & \text{if} \ \boldsymbol{M}(\boldsymbol{p})=1,\\
		\boldsymbol{I}(\boldsymbol{f}_{out})(\boldsymbol{p}), & \text{if}  \ \boldsymbol{M}(\boldsymbol{p})=0.
	\end{cases}  
\end{equation}

The value of $I(\boldsymbol{f}_{out})(\boldsymbol{p})$ is weighted average of all the neighboring pixels centered at $\boldsymbol{p}$ within a radius $r$:

\begin{equation}
	\boldsymbol{I}(\boldsymbol{f}_{out})(\boldsymbol{p}) = \frac{\sum_{\boldsymbol{s} \in \Psi(\boldsymbol{p})}\boldsymbol{W}_{(\boldsymbol{p},\boldsymbol{s})}\boldsymbol{f}_{out}(\boldsymbol{s})}{\sum_{\boldsymbol{s} \in \Psi(\boldsymbol{p})}\boldsymbol{W}_{(\boldsymbol{p},\boldsymbol{s})}}
\end{equation}

where $s$ indicates $p$'s neighboring pixels and $\Psi(\boldsymbol{p})=\lbrace \boldsymbol{s}| \Vert \boldsymbol{s} - \boldsymbol{p} \Vert_{\infty} \leqslant r, \boldsymbol{s} \ne \boldsymbol{p} \rbrace$, the neighborhood of $\boldsymbol{p}$. In UMC, we set radius $r=7$. $W_{(\boldsymbol{p},\boldsymbol{s})}$ is the weight assigned to point $\boldsymbol{s}$ for interpolating at $\boldsymbol{p}$, for which we utilize the RBF kernel, a distance-based exponential decaying weighting scheme:

\begin{equation}
	\boldsymbol{W}_{(\boldsymbol{p},\boldsymbol{s})} = \exp(-\lambda^{2}\Vert \boldsymbol{p} - \boldsymbol{s} \Vert^{2}_{2})
\end{equation}

with $\lambda$ being a trainable parameter. This indicates that the closer $\boldsymbol{s}$ is to $\boldsymbol{p}$, the larger its assigned weight will be. Note that masked-out features $\boldsymbol{M}(\boldsymbol{p})=0$ still participate in the interpolation process as inputs with values of $\boldsymbol{0}$.

\section{Detailed Qualitative results}

We visualize the detection results between different collaborative approaches on V2X-sim\cite{Li_2021_RAL} and OPV2V\cite{xu2022opencood} datasets, as shown in Figure~\ref{fig:big1} and \ref{fig:big2}. 

\section{Loss curve}

\begin{figure}[h]
	\centering
	\includegraphics[width=0.9\linewidth]{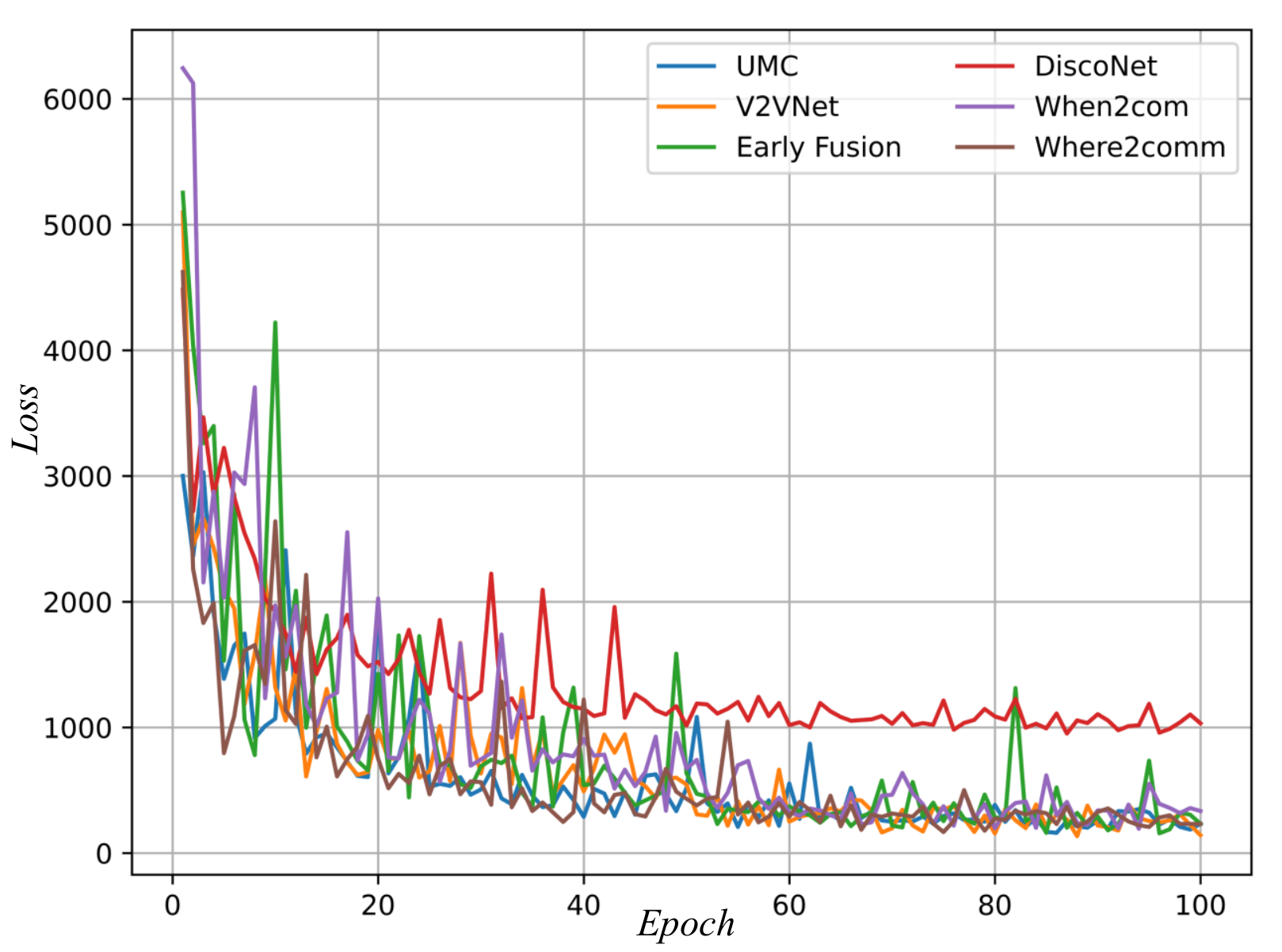}
	\caption{Epoch vs. loss on V2X-Sim dataset.}
	\label{fig:loss1}
\end{figure}

\begin{figure}[h]
	\centering
	\includegraphics[width=0.9\linewidth]{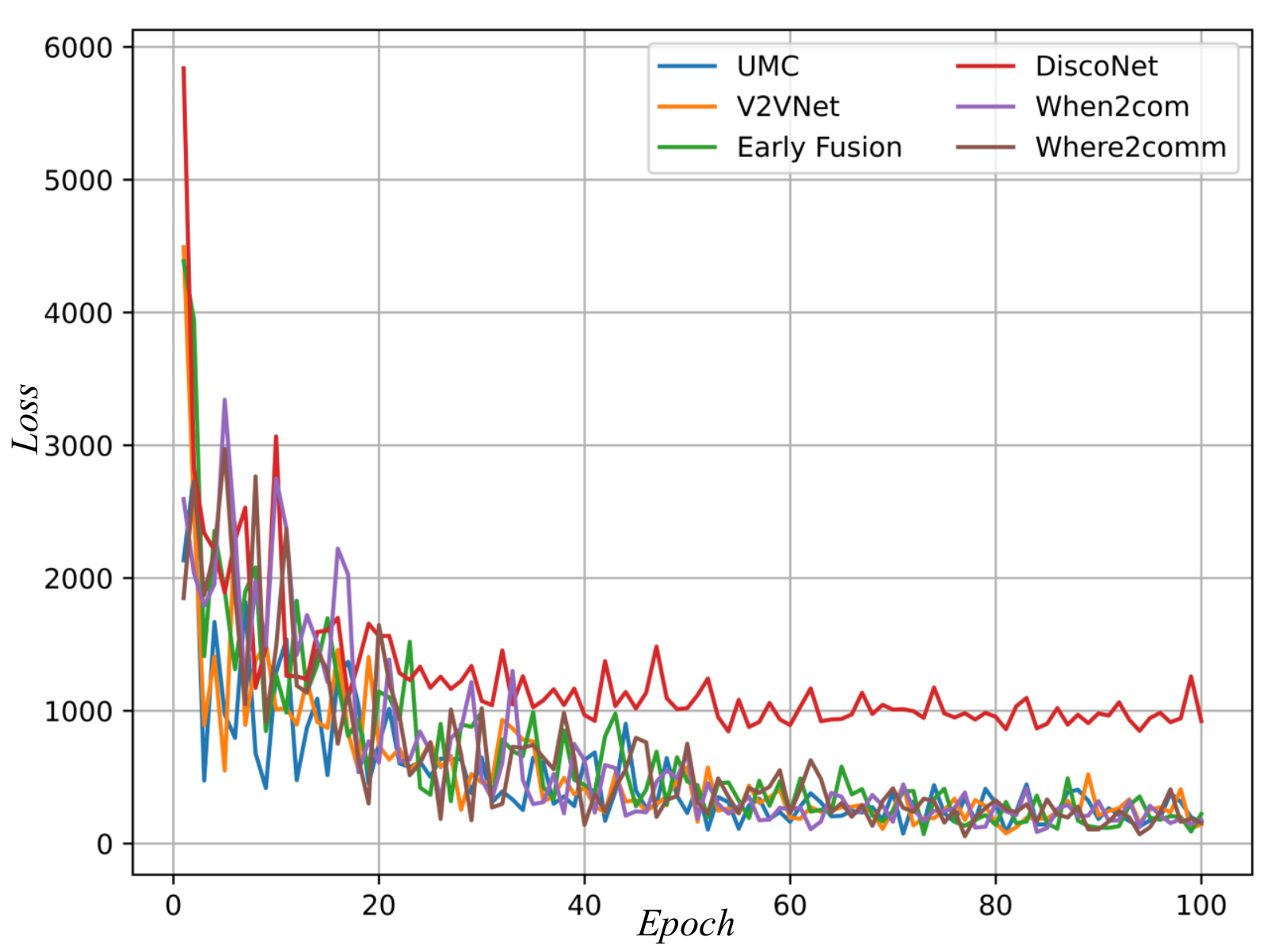}
	\caption{Epoch vs. loss on OPV2V dataset.}
	\label{fig:loss2}
\end{figure}

We visualize the loss curve of UMC, Early Fusion, When2com, Where2comm, V2VNet and DiscoNet on V2X-Sim and OPV2V on Figure~\ref{fig:loss1} and \ref{fig:loss2}, respectively.

\pdfoutput=1

\end{document}